\documentclass[]{article} 

\usepackage{amssymb, amsmath, amsthm} 
\usepackage{epsfig,graphicx} 
\usepackage{pdfsync}
\usepackage{caption}
\usepackage{subcaption}
\usepackage{bm}
\usepackage{verbatim}
\usepackage{multirow}
\usepackage{afterpage}
\usepackage{bbm}
\usepackage{color}
\usepackage{url}
\usepackage{authblk}

\title{Specific Differential Entropy Rate Estimation for Continuous-Valued Time Series}

 \author[1]{David Darmon}
 \affil[1]{Department of Military and Emergency Medicine\\Uniformed Services University\\Bethesda, MD 20814, USA}

\begin{document}

\maketitle

\begin{abstract}
	We introduce a method for quantifying the inherent unpredictability of a continuous-valued time series via an extension of the differential Shannon entropy rate. Our extension, the specific entropy rate, quantifies the amount of predictive uncertainty associated with a \emph{specific} state, rather than averaged over all states. We relate the specific entropy rate to popular `complexity' measures such as Approximate and Sample Entropies. We provide a data-driven approach for estimating the specific entropy rate of an observed time series. Finally, we consider three case studies of estimating specific entropy rate from synthetic and physiological data relevant to the analysis of heart rate variability.
\end{abstract}

\section{Introduction}

The analysis of time series resulting from complex systems must often be performed `blind': in many cases, mechanistic or phenomenological models are not available because of the inherent difficulty in formulating accurate models for complex systems. In this case, a typical analysis may assume that the data \emph{are} the model, and attempt to generalize from the data in hand to the system. For example, a common question to ask about a times series is how `complex' it is, where we place complex in quotes to emphasize the lack of a satisfactory definition of complexity at present~\cite{shalizi2006methods}. An answer is then sought that agrees with a particular intuition about what makes a system complex: for example, trajectories from periodic and entirely random systems appear simple, while trajectories from chaotic systems appear quite complicated. On a more practical level, it is common to compare two time series from similar systems, in which case one wants to meaningfully ask: is the phenomenon resulting from system A more or less complex than the phenomenon resulting from system B?  

There are many possible definitions of the complexity of a time series. See~\cite{Peliti:1988ci,shalizi2006methods} for comprehensive reviews. Some notable attempts at formal definitions include Kolmogorov complexity~\cite{li2013introduction}, stochastic complexity~\cite{rissanen1989stochastic}, forecast complexity~\cite{grassberger1986toward}, and Grassberger-Crutchfield-Young statistical complexity~\cite{crutchfield1989inferring}. Perhaps the most well-developed theory of complexity, which incorporates and expands on many of these quantities in the special case of discrete-valued time series, is computational mechanics~\cite{Shalizi:2001uf}. For example, see~\cite{James:2011ij} for an elucidation of the amount of information, in a formal sense, stored in a single observation from a discrete-valued stochastic process.

Practical definitions of complexity for continuous-valued time series are much less well-developed. The most common definitions rely on some notion of the difficulty in predicting a time series. There are currently at least two schools of thought for the (un)predictability-based notions of complexity when applied to systems with continuous states: Kolmogorov-Sinai entropy~\cite{kolmogorov1958new,sinaui1959concept} and Shannon entropy rate~\cite{Shannon1948}. Approaches based on the former treat the data as a trajectory from a deterministic dynamical system, and seek to estimate the Kolmogorov-Sinai entropy based on the trajectory~\cite{kantz2004nonlinear}. This school of thought goes back to some of the earliest work applying nonlinear dynamics to observational data~\cite{Crutchfield:1987va}. Approaches based on the latter treat the data as a realization from a stochastic process, and focus on entropy rate from a statistical perspective~\cite{lake2006renyi}. While these approaches \emph{seem} very similar, and are typically treated as such in much of the applied literature, they in fact give diverging answers to similar questions. In particular, the Kolmogorov-Sinai entropy of a stochastic dynamical system is infinite, while the differential Shannon entropy rate of a deterministic dynamical system is infinite~\cite{Ostruszka:2000jc}. These facts have been noted in some of the earliest work on estimating entropy rates from continuous-valued time series~\cite{Fraser:1989ib}, but are largely ignored in the applied literature. Moreover, methods proposed to estimate Kolmogorov-Sinai entropy may in fact be estimating Shannon entropy rate, and \emph{vice versa}. The situation may be further confused by the fact that the Kolmogorov-Sinai entropy \emph{does} correspond to a Shannon entropy rate, in this case the supremum over the \emph{discrete} Shannon entropy rates induced by finite partitions of the state space of a dynamical system~\cite{badii1999complexity}.

In addition to the methodological divide between the two dominant approaches to entropy rate estimation, neither has been used to provide a \emph{specific} entropy rate for the system as a function of its state. That is, estimates are typically reported as time averages which, under certain conditions, converge to state space averages. However, it may be desired to know the entropy rate associated with a system \emph{now}, at the present state, rather than on average. It is difficult to define such a state-specific entropy rate in the Kolmogorov-Sinai framework. For stochastic dynamics, such a state-specific entropy rate can be defined over \emph{ensembles} of the system starting at the specified state. Thus, one of the aims of this paper is to provide an estimator for such a \emph{specific} entropy rate.

The contributions of this paper are threefold. First, we reemphasize the dependence of the short term predictability of a nonlinear dynamical systems on its current state, and propose an information theoretic quantity, the specific entropy rate, that captures this dependence. Second, we propose a statistically principled approach to estimating the specific entropy rate from a continuous-valued time series that takes advantage of recent advances in conditional density estimation. Finally, we demonstrate the new approach with both synthetic and real data to highlight its strengths and weaknesses, with a special emphasis on interevent interval data as found in heart rate variability analysis. Throughout, we also make connections to modern practices in entropy rate estimation, both of the Kolmogorov-Sinai and differential schools, and seek to highlight how our estimator fits into those frameworks.
\section{Methodology}

In the following sections, we define the specific entropy rate of a stochastic dynamical system and develop an approach for its estimation from data. In Section~\ref{sec:sds}, we fix our notation and define a stochastic dynamical system. In Section~\ref{sec:er}, we review the entropy rate of a stochastic dynamical system, and define the specific entropy rate. In Sections~\ref{sec:kde} and~\ref{sec:model-selection}, we propose a method for estimating the specific entropy rate from finite data. Finally, in Section~\ref{sec:method-apen} we make connections between the specific entropy rate and other commonly used entropy rate estimators.

\subsection{Stochastic Dynamical System}

\label{sec:sds}

Consider an observed scalar real-valued time series $x_{1}, x_{2}, \ldots, x_{T}$. We explicitly model the time series as a realization from an autonomous stochastic dynamical system~\cite{fan2003nonlinear,chan2013chaos}. That is, unlike for autonomous deterministic dynamical systems which assume that a deterministic update rule acts on the precisely known state of the system, we assume that the states are stochastic, and moreover that transitions from state to state occur according to a transition density. Thus, we view $x_{1}, x_{2}, \ldots, x_{T}$, as a realization from the system $\{X_{t}\}_{t \in \mathbb{Z}}$, where we use the standard convention of using upper / lower case to denote a random variable / its realization. For $n > m$, let $X_{m}^{n} = (X_{m}, X_{m+1}, \ldots, X_{n-1}, X_{n})$ denote the $n - m + 1$ block of states for the dynamical system from time $m$ to time $n$. Similarly, let $X_{-\infty}^{m} = (\ldots, X_{m-1}, X_{m})$ denote the semi-infinite past until time $m$, and let $X_{n}^{\infty} = (X_{n}, X_{n + 1}, \ldots)$ denote the semi-infinite future starting at time $n$.  Then a general model~\cite{chan2013chaos} for how the state evolves assumes that the future state $X_{t}$ can be expressed as a random transformation of its past $X_{-\infty}^{t-1}$,
\begin{align}
X_{t} = F\left(X_{-\infty}^{t-1}; \epsilon_{t}\right) \label{eqn:update-eqn-sds}
\end{align}
where $\epsilon_{t}$ represents \emph{dynamical} noise, that is, noise that influences the \emph{dynamics} of the system, to be contrasted with \emph{observational} noise which impacts the observations of the system but not its dynamics. Equivalently, (\ref{eqn:update-eqn-sds}) can be expressed explicitly in terms of the transition density $f\left(x \mid x_{-\infty}^{t-1}\right)$ as
\begin{align}
X_{t} \sim f\left(x \mid X_{-\infty}^{t-1}\right). \label{eqn:transition-sds}
\end{align}
More typically, the dynamical noise is taken to be additive, in which case 
\begin{align}
	X_{t} = G(X_{-\infty}^{t-1}) + \epsilon_{t}
\end{align}
where typically $\{ \epsilon_{t}\}$ is taken to be independent and identically distributed and $\epsilon_{t}$ is taken to be independent of previous values of $X_{s}$, $s < t$. Finally, we note that we consider solely \emph{scalar} time series in this paper. While much of the theory can be translated to the case of multivariate time series by replacing the scalar observable $X_{t}$ with a $d$-dimensional vector observable $\mathbf{X}_{t}$, the impact of this change on the computational and statistical burdens of an approach such as the one we develop here are less easily overcome.

 \subsection{Differential Entropy Rate and Its Estimation}
 
 \label{sec:er}

Let $\{X_{t}\}_{t \in \mathbb{Z}}$ be a discrete-time, continuous-state stochastic dynamical system as defined in the previous section. Recall that for a continuous-valued random variable $X$ with density $f(x)$, the differential entropy~\cite{michalowicz2013handbook} of $X$ is given by
\begin{align}
	h[X] &= -E[\log f(X)] \\
	&= - \int_{\mathbb{R}} f(x) \log f(x) \, dx.
\end{align}
We will always take the logarithm with base $e$, and thus all differential entropies are in nats. For the remainder of this paper, because our focus is on continuous-state systems, when we use the term entropy, we refer to differential entropy. For random variables $(X, Y)$ with joint density $f(x, y)$, the joint entropy of $X$ and $Y$ is defined similarly as
\begin{align}
	h[X, Y] &= -E[\log f(X, Y)] \\
	&= - \int_{\mathbb{R}^{2}} f(x, y) \log f(x, y) \, dx \, dy. \label{eqn:joint-entropy}
\end{align}
Applying~(\ref{eqn:joint-entropy}) to a stochastic dynamical system $\{X_{t}\}_{t \in \mathbb{Z}}$ with a block-$p$ joint distribution $f_{t}$ at time $t$, the block-$p$ entropies at time $t$ are given by
\begin{align}
	h\left[X_{t}^{t+p-1}\right] = h[X_{t}, \ldots, X_{t + p - 1}] = -E[\log f_{t}(X_{t}, \ldots, X_{t + p - 1})].
\end{align}

There are two definitions of differential entropy rate which are equivalent for a strong-sense stationary stochastic process~\cite{cover2012elements,ihara1993information}. The first, which we denote as $\bar{h}_{1}(X)$, defines the entropy rate in terms of the rate of growth of block-$p$ entropies,
\begin{align}
	\bar{h}_{1}(X) = \lim_{t \to \infty} \frac{h[X_{1}, \ldots, X_{t}]}{t}. \label{eqn:growth-rate-er}
\end{align}
The second, which we denote as $\bar{h}_{2}(X)$, defines entropy rate in terms of the entropy of a one-step-ahead future conditional on a sufficiently long past,
\begin{align}
	\bar{h}_{2}(X) = \lim_{t \to \infty} h\left[X_{t+1} \mid X_{1}^{t}\right]. \label{eqn:conditional-er}
\end{align}
While these are equivalent for strictly stationary stochastic processes, they need not be for an arbitrary process. Because we are interested in quantifying the \emph{predictability} of a stochastic process over time, we take~(\ref{eqn:conditional-er}) as our definition of entropy rate, $\bar{h}(X) \equiv \bar{h}_{2}(X)$.

Clearly, care must be taken when interpreting the densities that appear in the definitions of entropies and entropy rates we have defined thus far, and this interpretation depends on the assumptions that the practitioner is willing or able to make about the system under consideration. In practice, the assumption is typically made that $\{X_{t}\}_{t \in \mathbb{Z}}$ is strong-sense stationary~\cite{grimmett2001probability}, or at least can be made so via transformations such as differencing or detrending. These assumptions are typically violated in practice. We make a less restrictive assumption on the system under consideration, namely that it is \emph{conditionally stationary}~\cite{caires2005non}. A process is conditionally stationary if the conditional distribution function of $X_{t+1}$ given $(X_{t}, \ldots, X_{t - p + 1}) = \mathbf{x}$ does not depend on $t$ for some fixed $p$: that is, the statistical future of the process conditional on a past of sufficient length does not depend on \emph{when} that past was observed. Strong-sense stationary processes and Markov processes are special cases of this type.

The value of $\bar{h}(X)$ depends on $h\left[X_{t + 1} \mid X_{1}^{t}\right]$ and thus on the conditional structure of the stochastic process. Consider the conditional entropy of $X_{t}$ given the block $X_{t-p}^{t-1}$ of length $p$. Under the assumption of conditional stationarity of order $p$, this conditional entropy can be rewritten as
\begin{align}
	h\left[X_{t} \mid X_{t-p}^{t-1}\right] &= -E[\log f_{t}(X_{t} \mid X_{t-p}^{t-1})] \\
	&= - \int_{\mathbb{R}^{p+1}} f_{t}(x_{1}^{p+1}) \log f_{t}(x_{p+1} \mid x_{1}^{p}) \, dx_{p+1} \, d x_{1}^{p} \\
	&= - \int_{\mathbb{R}^{p+1}} f_{t}(x_{1}^{p}) f_{t}(x_{p+1} \mid x_{1}^{p})  \log f_{t}(x_{p+1} \mid x_{1}^{p}) \, dx_{p+1} \, d x_{1}^{p}  \label{eqn:ER-line1} \\
	&= - \int_{\mathbb{R}^{p+1}} f_{t}(x_{1}^{p}) f(x_{p+1} \mid x_{1}^{p})  \log f(x_{p+1} \mid x_{1}^{p}) \, dx_{p+1} \, d x_{1}^{p}  \label{eqn:ER-line2}\\
	&= - \int_{\mathbb{R}^{p}} f_{t}(x_{1}^{p}) E\left[\log f(X_{t} \mid X_{t-p}^{t-1}) \mid X_{t-p}^{t-1} = x_{1}^{p} \right] d x_{1}^{p} \label{eqn:er-line5}\\
	&= - E\left[ E\left[\log f(X_{t} \mid X_{t-p}^{t-1}) \mid X_{t-p}^{t-1}\right]\right]\label{eqn:er-line6}.
\end{align}
where going from (\ref{eqn:ER-line1}) to (\ref{eqn:ER-line2}) we have applied conditional stationarity. Thus, we see that the order $p$ conditional entropy depends on two properties of the stochastic process: the state-specific entropy conditional on a particular past $x_{1}^{p}$, and the overall density of the pasts $X_{1}^{p}$. This decomposition motivates defining the state-specific entropy rate of order $p$ at time $t$ as
\begin{align}
	h^{(p)}_{t} &\equiv h[X_{t} \mid X_{t-p}^{t-1} = x_{t-p}^{t-1}] \\
	&= - E\left[\log f(X_{t} \mid X_{t-p}^{t-1}) \mid X_{t-p}^{t-1} = x_{t-p}^{t-1}\right] \\
	&= - \int_{\mathbb{R}}  f(x_{p+1} \mid x_{1}^{p})  \log f(x_{p+1} \mid x_{1}^{p}) \, dx_{p+1} \label{eqn:inst-er}.
\end{align}
We will call $h^{(p)}_{t}$ the \emph{specific entropy rate} of order $p$, or simply the \emph{specific entropy rate} where the order $p$ is clear. We will specify a procedure for choosing $p$ in Section~\ref{sec:model-selection}. The specific entropy rate quantifies the unpredictability of the process conditional on the specific past $x_{t-p}^{t-1}$ observed immediately before time $t$. We see that~(\ref{eqn:inst-er}) emphasizes the well-known fact that the \emph{difficulty in prediction} can depend on the current state for both deterministic and stochastic nonlinear dynamics~\cite{yao1994quantifying,yao1994prediction}. This is \emph{not} the case for linear time series models, where the specific entropy rate is independent of the present state of the system. We note that our specific entropy rate is similar in spirit to the specific information of a stimulus~\cite{deweese1999measure} from computational neuroscience, local information measures from~\cite{Lizier:2008ei,lizier2014measuring}, and the Lyapunov-like index~\cite{yao1994quantifying} from statistical nonlinear time series analysis. The specific information of a stimulus notes that the mutual information between two random variables $R$ and $S$ can be decomposed as \hbox{$I[R \wedge S] = H[R] - H[R \mid S]$} where $H$ denotes the \emph{discrete} Shannon entropy. Thus, the specific information of a particular stimulus $s$ for a response $R$ is taken to be $I[R \wedge s] = H[R] - H[R \mid S = s]$, using a similar decomposition as (\ref{eqn:er-line6}). The local information measures go one further step back, defining the local information measures in terms of the argument of the expectation associated with the information measure. For example, the local entropy rate of order $p$ at $x_{1}^{p+1}$ under this formalism is defined as $-\log f(x_{p+1} \mid x_{1}^{p})$, rather than as $-E[\log f(X_{p+1} \mid X_{1}^{p}) \mid X_{1}^{p} = x_{1}^{p}]$ in our definition. The Lyapunov-like index is defined in terms of divergences with respect to the past of conditional expectations of the future given the past, and thus measures uncertainty about the future given the past using solely the first moment of the predictive density. 

In practice, the predictive density $f(x_{p+1} \mid x_{1}^{p})$ is unknown and must be inferred from observations of the system under consideration. Thus, we consider the plug-in estimator for the specific entropy rate, namely
\begin{align}
	\hat{h}^{(p)}_{t} \equiv - E\left[\log \hat{f}(X_{t} \mid X_{t-p}^{t-1}) \mid X_{t-p}^{t-1} = x_{t-p}^{t-1}\right] \label{eqn:final-er-estimator}
\end{align}
where we substitute an estimator $\hat{f}(x_{p+1} \mid x_{1}^{p})$ for the unknown predictive density $f(x_{p+1} \mid x_{1}^{p})$. Finally, if an estimator for the overall entropy rate~(\ref{eqn:conditional-er}) of the system is desired, we define the estimator
\begin{align}
	\hat{\bar{h}}^{(p)} &= \frac{1}{T-p} \sum_{t = p+1}^{T}- E\left[\log \hat{f}(X_{t} \mid X_{t-p}^{t-1}) \mid X_{t-p}^{t-1} = X_{t-p}^{t-1}\right] \\
	&= \frac{1}{T-p} \sum_{t = p+1}^{T} \hat{h}^{(p)}_{t},
\end{align}
a time-average of the specific entropy rates, using the empirical distribution over the pasts as an estimator for $f_{t}(x_{1}^{p})$ in~(\ref{eqn:er-line5}).

Before considering the problem of estimating the predictive density \hbox{$\hat{f}(x_{p+1} \mid x_{1}^{p})$}, we note that we are really interested in the specific entropy of the predictive density and not the predictive density outright. Thus, the predictive density $f(x_{p+1} \mid x_{1}^{p})$ is a nuisance parameter, and a difficult one to estimate especially in higher dimensions. Based on this insight, many information theoretic estimators has been proposed that directly estimate the quantity of interest without first estimating the underlying density. For example, many estimators have been proposed based on the statistics of $k$-nearest neighbors amongst the sample points~\cite{kozachenko1987sample,Kraskov:2004gr,Sricharan:2013ip,Gao:2015vj,Singh:2016vk,Lombardi:2016fw}. In fact, many of these estimators correspond to plug-in estimators using \emph{variable} bandwidth kernel density estimators~\cite{Terrell:1992}, with the bandwidth varying with the evaluation point: the bandwidth is taken to be the distance to the $k^{\text{th}}$ nearest neighbor. A key aspect of our estimator, which we turn to in Section~\ref{sec:model-selection}, is the use of model selection to directly learn which lags are relevant to prediction. A similar approach could be taken with the $k^{\text{th}}$ nearest neighbor-based estimators, letting $k$ vary with each lag. We return to a discussion of this approach, and its relation to our method, in Section~\ref{sec:discussion}.

\subsection{Conditional Density Estimation}

\label{sec:kde}

The problem of estimating a conditional density goes back to the pioneering work of Rosenblatt~\cite{rosenblatt1969conditional}. We estimate the predictive density using the conditional kernel density estimator proposed in~\cite{hall2004cross,hayfield2008nonparametric}. See~\cite{bosq2012nonparametric} for additional theoretical results for density estimators for general stochastic processes. Consider a continuous-valued time series $\{ X_{t}\}_{t = 1}^{T}$ for which we desire to estimate the predictive density $f(x_{p+1} \mid x_{1}^{p})$. Recalling that the predictive density is given by
\begin{align}
	f(x_{p+1} \mid x_{1}^{p}) = \frac{f(x_{1}^{p}, x_{p+1})}{f(x_{1}^{p})},
\end{align}
we can estimate the predictive density by estimating the joint density $f(x_{1}^{p}, x_{p+1})$  and the marginal density $f(x_{1}^{p})$ and taking their ratio. We estimate the marginal and joint densities using the kernel density estimators
\begin{align}
	\hat{f}(x_{1}^{p}) = \frac{1}{T-p} \sum_{t = p+1}^{T} K_{\mathbf{k}}(x_{1}^{p}, X_{t-p}^{t-1})
\end{align}
and
\begin{align}
	\hat{f}(x_{1}^{p}, x_{p+1}) = \frac{1}{T - p} \sum_{t = p+1}^{T} K_{\mathbf{k}}(x_{1}^{p}, X_{t-p}^{t-1}) L_{k_{p + 1}}(x^{p+1}, X_{t}),
\end{align}
respectively, where $K_{\mathbf{k}}$ is the product kernel
\begin{align}
	K_{\mathbf{k}}(x_{1}^{p}, X_{t-p}^{t-1}) = \prod_{j = 1}^{p} \frac{1}{k_{j}} K\left( \frac{x_{j} - X_{t - p  + j - 1}}{k_{j}} \right),
\end{align}
$L_{k_{p+1}}$ is the univariate kernel
\begin{align}
	L_{k_{p+1}}(x_{p+1}, X_{t}) = \frac{1}{k_{p+1}} K\left( \frac{x_{p+1} - X_{t}}{k_{p+1}} \right),
\end{align}
$k_{1}, \ldots, k_{p+1}$ are the bandwidths, and $K(\cdot)$ is a kernel function, \emph{i.e.} a positive, symmetric probability density with finite second moment. The estimator for the conditional density $\hat{f}(x_{p+1} \mid x_{1}^{p})$ is then
\begin{align}
	\hat{f}(x_{p+1} \mid x_{1}^{p}) = \frac{\hat{f}(x_{1}^{p}, x_{p+1})}{\hat{f}(x_{1}^{p})}. \label{eqn:fhat}
\end{align}

Note that the joint and marginal density estimators are \emph{coupled} since they use the same bandwidths $k_{1}, \ldots, k_{p}$ for both the marginal and joint density estimators. This coupling is necessary to ensure that, for example, the conditional density integrates to one with respect to $x_{p+1}$. On a more practical level for time series, this coupling allows us to screen out the distant past. Consider, for example, the extreme case where the past is irrelevant to the future in terms of prediction. By this coupling, we can ignore the past by setting the bandwidths $k_{1}, \ldots, k_{p}$ to large values. This has the effect of giving $\hat{f}(x_{p+1} \mid x_{1}^{p}) \approx \hat{f}(x_{p+1})$ and recovering the appropriate independence relationship. More generally, if $q < p$ lags are sufficient to screen off the distant past, then by setting the bandwidths $k_{1}, \ldots, k_{p-q}$ sufficiently large we can recover $\hat{f}(x_{p+1} \mid x_{1}^{p}) \approx \hat{f}(x_{p+1} \mid x_{p - q + 1}^{p}).$ We discuss how to take advantage of this property of conditional kernel density estimators in more detail in the next section.

\subsection{Bandwidth and Order Selection}

\label{sec:model-selection}

The estimator of the conditional density function (\ref{eqn:fhat}), and thus the estimator of the specific entropy rate (\ref{eqn:final-er-estimator}), depends on the choice of the order $p$ and bandwidths $k_{1}, \ldots, k_{p+1}$. We therefore require a principled and repeatable procedure for selecting them. For example, in the context of transfer entropy estimation, \cite{kaiser2002information} noted how, depending on the choice of these parameters, the direction of causality can be reversed. Because our approach explicitly builds a \emph{statistical} model for the dynamical system, we choose the order and bandwidths via $l$-block cross-validation~\cite{burman1994cross} of the negative log-likelihood of the conditional density. (Note that~\cite{burman1994cross} calls their method $h$-block cross-validation, which we rename in this manuscript to avoid confusion with differential entropy.) $l$-block cross-validation is an extension of leave-one-out cross-validation where instead of leaving out a \emph{single} observation at each evaluation, we remove the observation and $l$ observations on either side of that observation. That is, we seek the values of $p$ and $\mathbf{k} = (k_{1}, \ldots, k_{p+1})$ that minimize
\begin{align}
	\text{CV}_{l}(p, \mathbf{k}) = -\frac{1}{T-p} \sum_{t = p+1}^{T} \log \hat{f}_{-t:l}(X_{t} \mid X_{t-p}^{t-1}), \label{eqn:cv-function}
\end{align}
where $\hat{f}_{-t:l}$ is the estimate of conditional density after removing the $2l + 1$ observations about $t$. This accounts for a bias in 0-block cross-validated likelihood resulting from the dependence inherent in temporally nearby realizations of a time series. We immediately see that~(\ref{eqn:cv-function}) takes the form of an entropy rate, so this cross-validation procedure can also be thought of as minimizing the entropy rate of the model. Thus, cross-validation provides a principled means for choosing the order of the entropy rate in analogy to common practices in the discrete-valued case. For example, when computing the entropy rate for discrete-valued data, it is frequently recommended to choose the order of the entropy rate by searching for an asymptotic value for order-$p$ entropy rate as a function of $p$~\cite{crutchfield2003regularities}. Thus, our approach extends this heuristic to the continuous-valued case, with an additional penalty on $p$ induced by the nature of cross-validation. Moreover, both theoretical and empirical work have shown that choosing the bandwidth via cross-validation can automatically `smooth out' irrelevant predictors by setting their bandwidths very large~\cite{hall2004cross,Efromovich:2010kp}. This is clearly desirable in the time series case, since we expect to induce conditional independence between the distant past and the future after accounting for a sufficient portion of the recent past. By using cross-validation, we get this dimension reduction for free.

Because of the computationally intensive nature of $l$-block cross-validation, we begin by fixing $p$ and choosing the bandwidths $(k_{1}, \ldots, k_{p+1})$ using $0$-block cross-validation, which reduces to leave-one-out-cross-validation. Then, using these bandwidths, we choose $p$ via $l$-block cross-validation. In all of the reported results, we use $l = 50$, thus leaving out 101 points about any evaluation in~(\ref{eqn:cv-function}). In principle, the block size could be chosen using the autocorrelation time or lagged mutual information~\cite{kantz2004nonlinear}, or a data-driven approach~\cite{lahiri2013resampling}. We leave the exploration of these approaches for future work.

\subsection{Relationship to Other Entropy Rate Estimators}

\label{sec:method-apen}



In the nonlinear dynamics community, especially in applications to biological systems, two popular measures of the uncertainty associated with the dynamics of a system are Approximate Entropy~\cite{pincus1991approximate} and Sample Entropy~\cite{richman2000physiological}. Despite their names, both of these quantities correspond to estimators of entropy \emph{rates} rather than entropies. Approximate Entropy,  as originally proposed by Pincus, was motivated by a finite-time, finite-resolution approximation to the Kolmogorov-Sinai Entropy of a deterministic dynamical system. The Sample Entropy was proposed as a modification to the Approximate Entropy that addressed several of its deficiencies. In~\cite{lake2006renyi}, Lake elucidates the key connection between the Approximate and Sample Entropies and information theoretic  entropy rates. In particular, Lake shows that the Approximate Entropy corresponds to a kernel density-based estimator of the Shannon differential entropy rate using uniform kernels and fixed bandwidths $k_{1} = k_{2} = \ldots = k_{p+1}$, while the Sample Entropy corresponds to a kernel density-based estimator of the R\'{e}nyi entropy rate with order $\alpha = 2$, the so-called collision entropy, with a particular choice of definition for the conditional R\'{e}nyi entropy. (Unlike conditional Shannon entropy, no standard definition of conditional R\'{e}nyi entropy exists for arbitrary $\alpha$~\cite{teixeira2012conditional}.)
In later work, recommendations were made for choosing the model order $p$~\cite{Lake:2002ea}, for setting the common bandwidth~\cite{lake2009nonparametric}, and for incorporating an adaptive bandwidth~\cite{Lake:fl}.
In the Appendix to this paper, we reproduce the derivation made in~\cite{lake2006renyi} connecting the  Approximate Entropy statistic to kernel density-based estimators of the differential entropy rate.

\section{Results}

We consider entropy rate estimation in three examples of increasing realism. The first example, described in Section~\ref{sec:mm}, applies the specific entropy rate estimator to a second-order Markov model. This example was designed to emphasize, in a simple way, the potential dependence of the specific entropy rate $h_{t}$ on the state of the system. In Section~\ref{sec:chaotic-ii}, we consider the entropy rate of interevent intervals resulting from an integrate-and-fire model driven by synthetic chaotic signals. This type of model is typically implicit in many of the analyses of biological signals ranging from heart rate variability  to neural firing. This example demonstrates how our entropy rate estimator performs when the assumption of a \emph{stochastic} dynamical system is violated. Finally, in Section~\ref{sec:hrv}, we demonstrate the specific entropy rate estimator using interbeat interval sequences resulting from a tilt table experiment.

Throughout these examples, we use the R package \texttt{np}~\cite{hayfield2008nonparametric} to estimate the conditional densities using second-order Gaussian kernels~\cite{Wand:1990cf}. As recommended in the methodology section, for a particular model order $p$, we choose the bandwidths using leave-one-out cross-validation on the log-likelihood, and choose the model order $p$ using $l$-block cross-validation with $l = 50$. We then estimate the specific entropy rate using (\ref{eqn:final-er-estimator}).

\subsection{A Second-Order Markov Process}

\label{sec:mm}

\begin{figure}[!ht]
\includegraphics[width=1\textwidth]{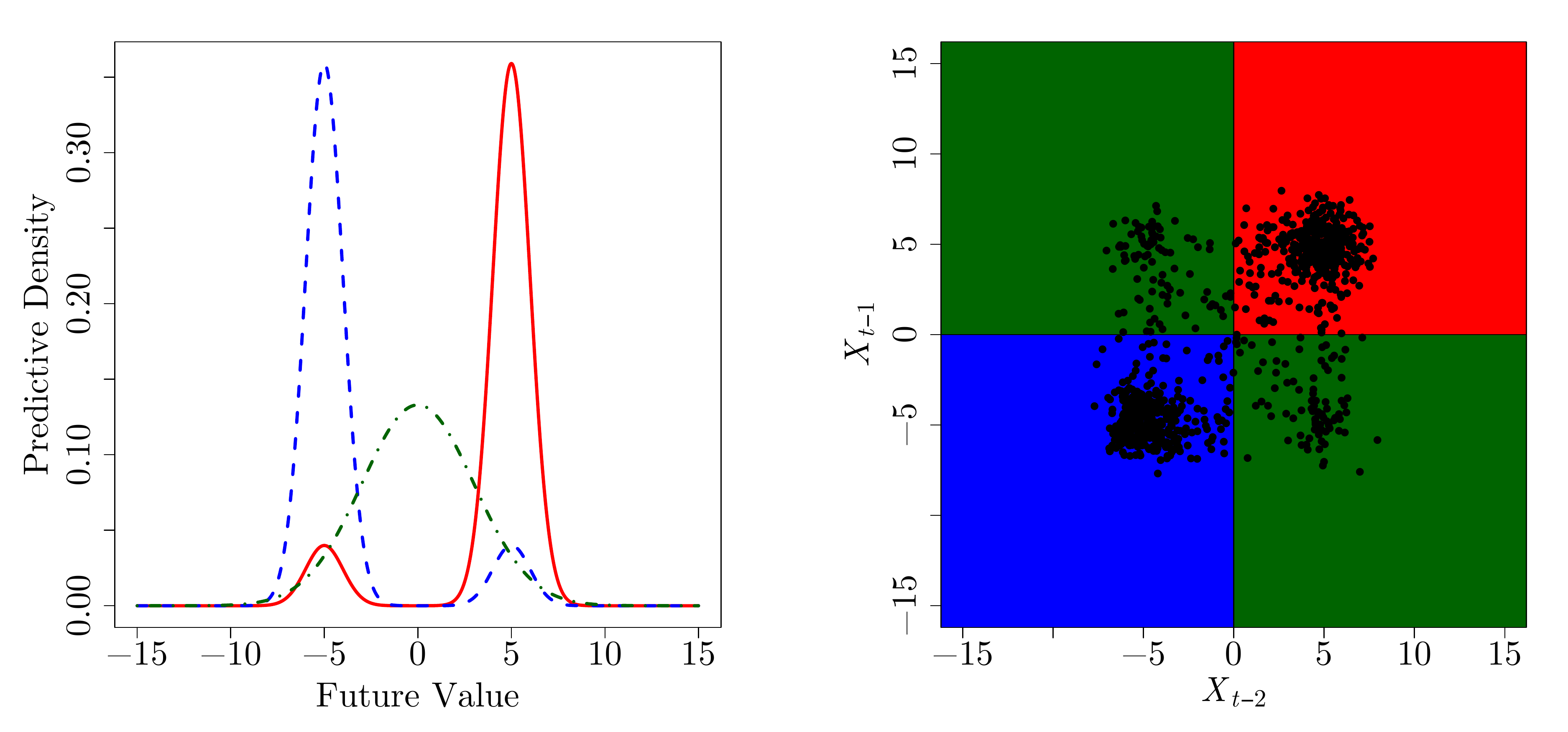}
\caption{Left: The predictive densities associated with each of the effective states for the Markov process~(\ref{eqn:transition-kernel}). Right: A scatter plot representation of the marginal density of $(X_{t-2}, X_{t-1})$ with the effective states colored according to the convention in the left panel.}
\label{fig:somp}
\end{figure}

Our first example is chosen to highlight the state-dependent nature of the specific entropy rate~(\ref{eqn:inst-er}). We consider a stochastic dynamical system with three effective states. One of the states corresponds to a \emph{crossing event}, when the system switches from positive to negative outputs or vice versa. This state has a high specific entropy rate. The other two states correspond to when the system settles into either a run of positive outputs or a run of negative outputs. In these states, the specific entropy rate is smaller. Explicitly, consider the second-order Markov process with the transition density
\begin{align}
	\label{eqn:transition-kernel}
	f(x_{t} \mid x_{t-2}, x_{t-1}) &= 
	\left\{\begin{array}{cl}
	p_{+} \phi(x_{t}; 5, 1) + (1 - p_{+}) \phi(x_{t}; -5, 1) &: x_{t-2}, x_{t-1} > 0\\
	p_{-} \phi(x_{t}; -5, 1) + (1 - p_{-}) \phi(x_{t}; 5, 1) &: x_{t-2}, x_{t-1} < 0 \\
	\phi(x_{t}; 0, 3^{2}) &: \text{ otherwise}
	\end{array}\right.
\end{align}
where $p_{+} = p_{-} = 0.1$, and $\phi(x; \mu, \sigma^{2})$ is the probability density function for a normal random variable with mean $\mu$ and variance $\sigma^{2}$,
\begin{align}
	\phi(x; \mu, \sigma^{2}) = \frac{1}{\sqrt{2 \pi \sigma^{2}}} e^{-\frac{1}{2 \sigma^{2}}(x - \mu)^2}.
\end{align}
The transition densities for each effective state are shown in the left panel of Figure~\ref{fig:somp}. The first effective state (red solid) corresponds to when the two previous observations were positive, the second effective state (blue dashed) corresponds to when the two previous observations were negative, and the third effective state (green dash-dotted) corresponds to when the two previous observations had opposite signs. The right panel of Figure~\ref{fig:somp} shows a scatter plot representation of the marginal density $(X_{t}, X_{t+1})$ with the quadrants colored by the effective states.

\begin{figure}[!ht]
        \centering
                \includegraphics[width=1.0\textwidth]{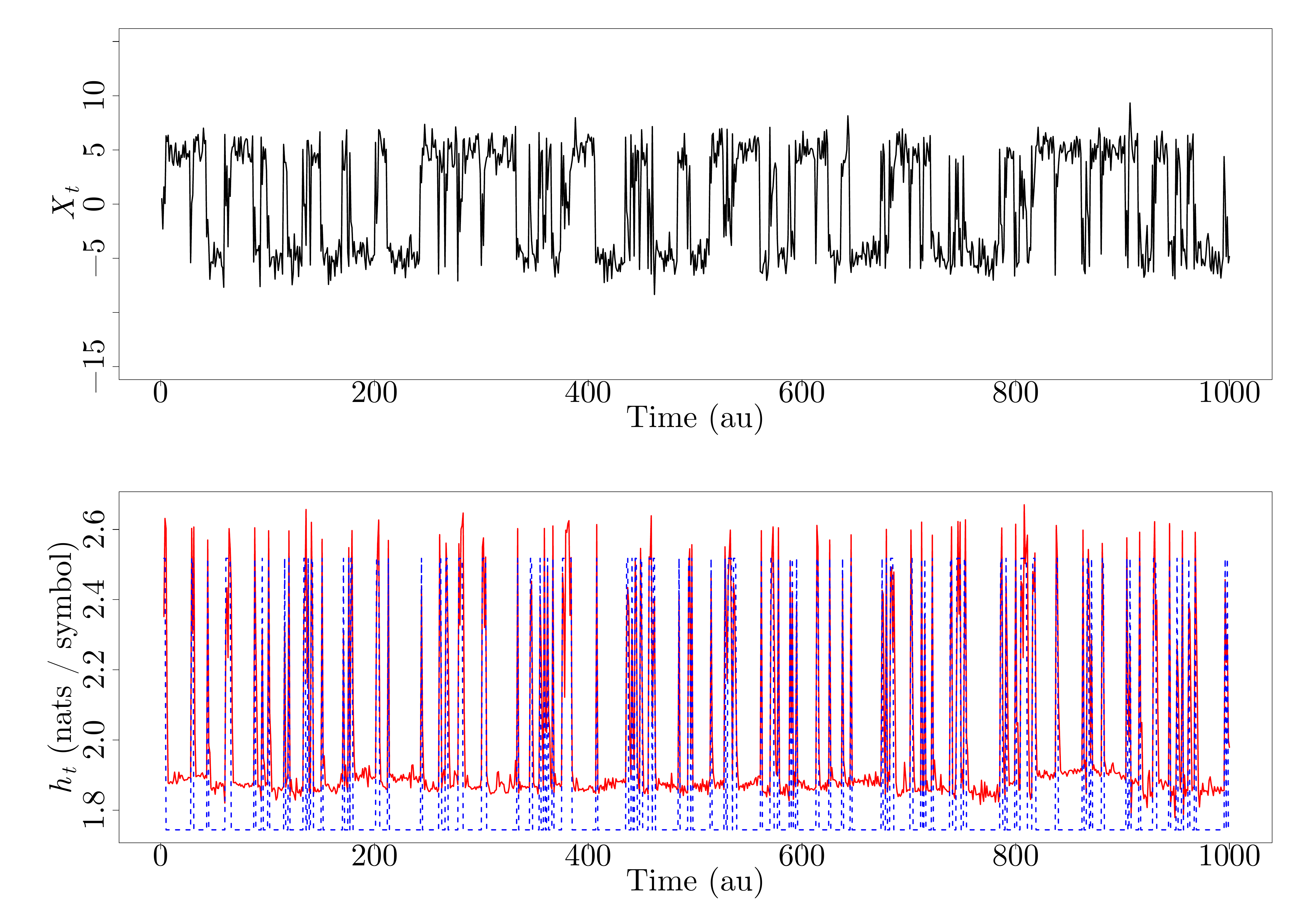}
        \caption{An example realization from~(\ref{eqn:transition-kernel}) (top), along with the specific entropy rate (bottom). The dashed blue line indicates the true specific entropy rate, while the solid red line indicates the entropy rate estimated using~(\ref{eqn:final-er-estimator}).}
		\label{fig:somp-ts+er}
\end{figure}

The top panel of Figure~\ref{fig:somp-ts+er} shows an example realization with $T = 1000$ which we use to estimate the specific entropy rate. We can compute the specific entropy rate $h_{t}$ for each effective state exactly. By symmetry, the first two effective states have the same specific entropy rate, which we compute by evaluating~(\ref{eqn:inst-er}) numerically: 1.744 nats per symbol. The third effective state's predictive density corresponds to a normal density with variance $9$, and thus has specific entropy rate $\frac{1}{2} \log (2 \pi e \cdot 3^{2}) \approx 2.518$ nats per symbol. The bottom panel shows the specific entropy rate (dashed blue), along with the estimated specific entropy rate with $p = 2$ (solid red). From the specific entropy rate, we can clearly see when the system switches from one of the low specific entropy rate states to the high specific entropy rate state, and vice versa. Moreover, we see that the estimated entropy also displays these transitions, though not as cleanly.

\begin{figure}[!ht]
\centering
\includegraphics[width=1\textwidth]{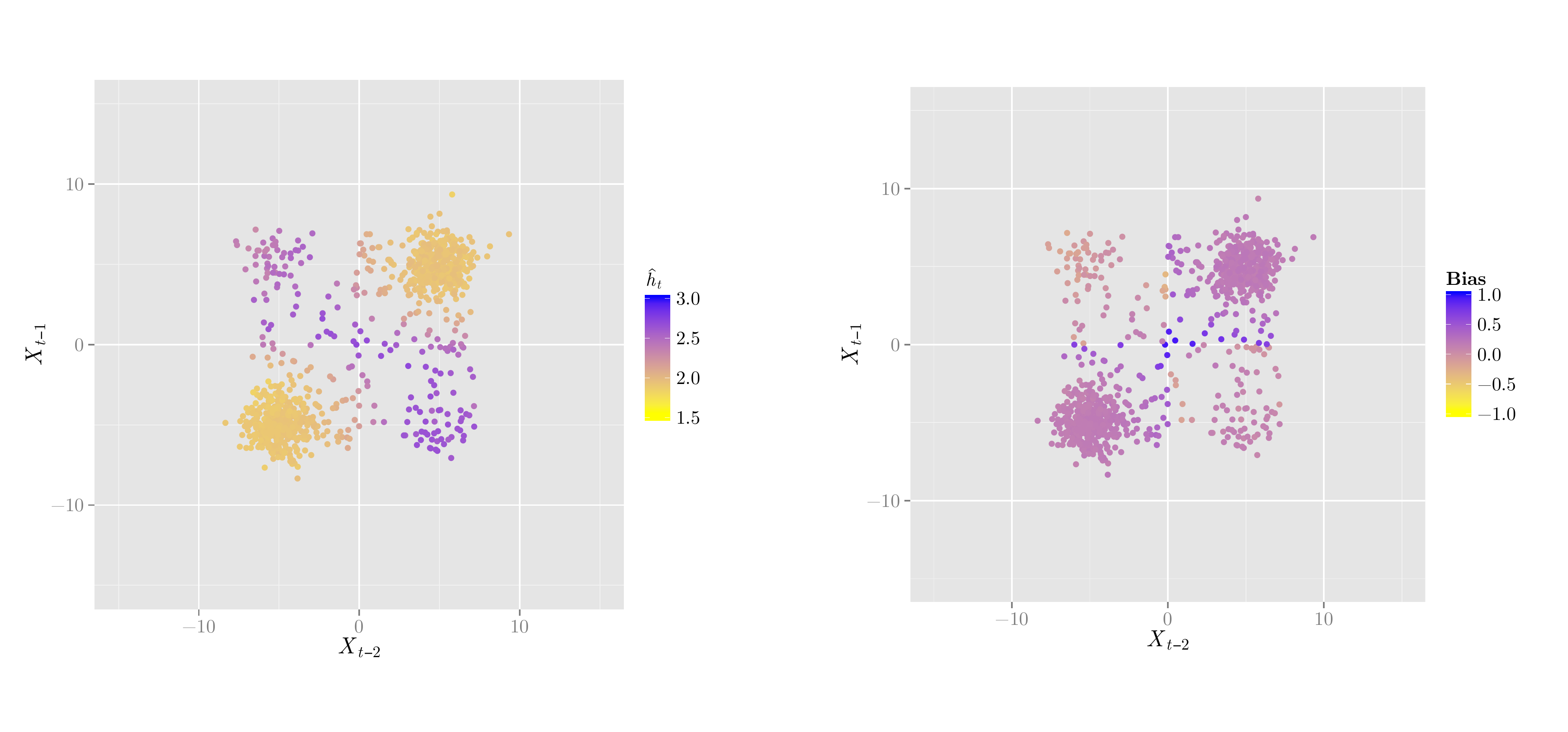}
\caption{The estimated specific entropy rate $\hat{h}_{t}$ (left) and its bias $\hat{h}_{t} - h_{t}$ (right) as a function of the history $(X_{t-2}, X_{t-1})$ for the Markov model. Note that the estimator correctly identifies the high and low specific entropy rate histories, and its largest bias occurs near the transitions between quadrants.}
\label{fig:somp-er+bias}
\end{figure}

To see the performance of the estimator as a function of the history, for each time point $t$ we compute both the estimator of the specific entropy rate $\hat{h}_{t}$, as well as the empirical bias between the estimated and true value,
\begin{align}
	\text{Bias}\left(\hat{h}_{t}\right) = \hat{h}_{t} - h_{t}.
\end{align}
Figure~\ref{fig:somp-er+bias} displays the estimated specific entropy rate (left) and bias (right) as a function of the history $(x_{t-2}, x_{t-1})$. As we saw in Figure~\ref{fig:somp-ts+er}, the estimator successfully distinguishes between the high entropy rate effective state (colored purple) and the low entropy rate effective states (colored yellow). Because the estimated specific entropy rate is always positive for this system, a positive bias indicates that the estimated entropy rate is \emph{larger} (greater predictive uncertainty) than it should be, and a negative bias indicates that the entropy rate is \emph{smaller} (lower predictive uncertainty) than it should be. We see that a large positive bias occurs for those pasts that belong to either the first (red) or second (blue) effective states, but lie near the border with the third (green) effective state. This occurs because of the discontinuous transition in the predictive density between each state. It is especially pronounced for those (rare) pasts near the origin, again because of the discontinuity.


\begin{figure}[!ht]
\includegraphics[width=1\textwidth]{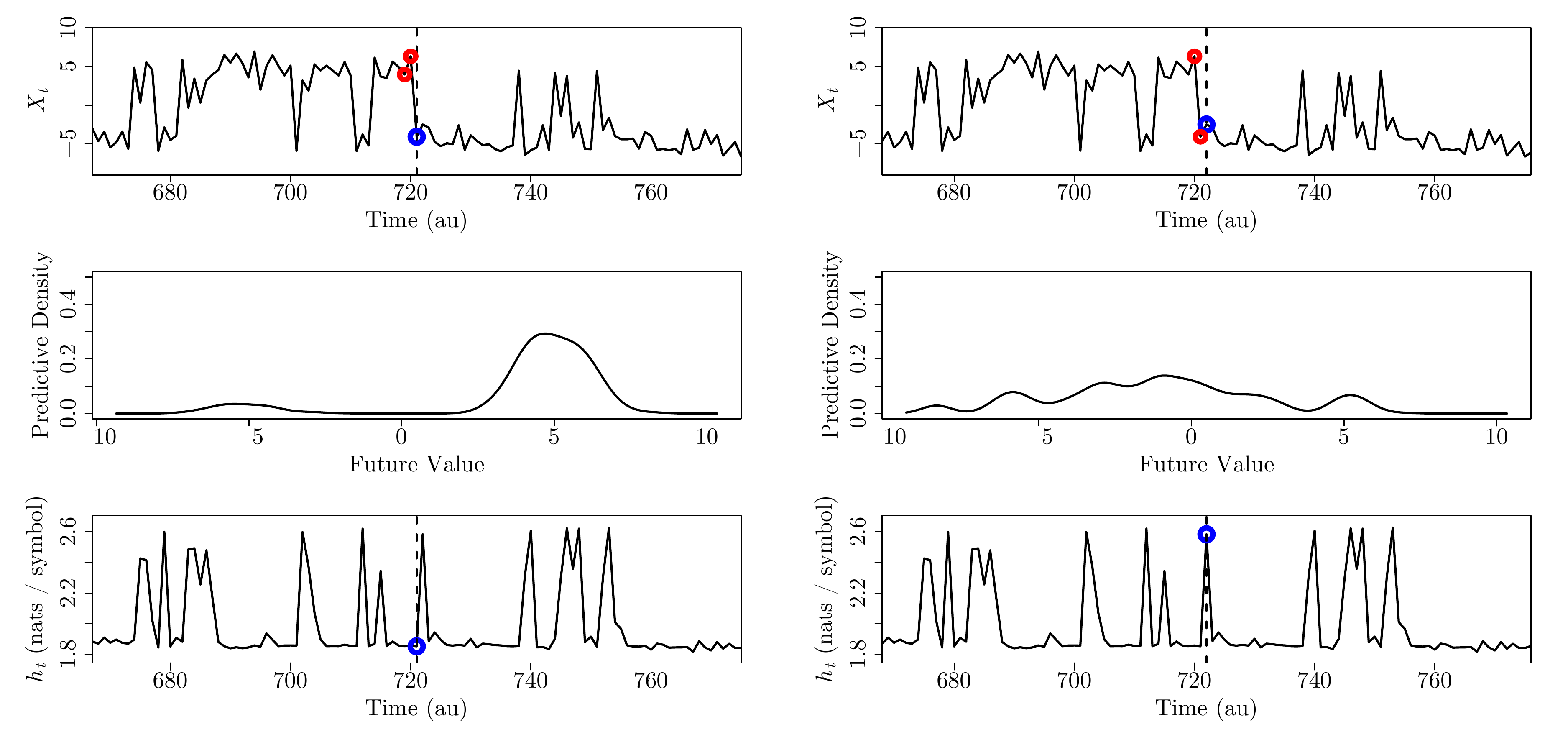}
\caption{A demonstration at two adjacent time points of (top) a realization from the second order Markov model, (middle) the estimated predictive density $\hat{f}(x_{t} \mid x_{t-2}, x_{t-1})$, and (bottom) the specific entropy rate for the second-order Markov process in low (left) and high (right) specific entropy rate states. In the top panels, the dashed vertical bar indicates the time $t$, the red points correspond to the specific pasts $(x_{t - 2}, x_{t-1})$, and the blue points correspond to the future values $x_{t}$.}
\label{fig:schematic-mm}
\end{figure}

Finally, we demonstrate two snap shots of the system in Figure~\ref{fig:schematic-mm} to recall the intuition behind the specific entropy rate, and how it relates to the predictive density of the stochastic dynamical system. Each panel shows the state of the system (top) with the present state $x_{t}$ marked by a blue circle and the past $(x_{t-2}, x_{t-1})$ marked by red circles, the estimated predictive density $\hat{f}(\cdot \mid x_{t - 2}, x_{t - 1})$ (middle), and the estimated specific entropy rate (bottom). The left panel corresponds to when the two past observations were positive, and thus the system is in one of the low entropy rate effective states. The right panel corresponds to when the two past observations were opposite in sign, and thus the system is in the high entropy rate effective state. We see that in both cases, the estimated predictive densities and estimated entropy rates agree with the effective states.  

\subsection{Interevent Intervals from an Integrate-and-Fire Model Driven by Chaotic Signals}

\label{sec:chaotic-ii}

For our second example, we consider interevent intervals resulting from an integrate-and-fire model driven by a chaotic signal. This model implicitly motivates many of the embedding-based analyses used with neural and heart rate variability data. For example, it is common to consider the times between heart beats  (interbeat intervals or RR intervals) as if they are equispaced samples from a continuous time process, and then apply methods from nonlinear dynamics. There is not, \emph{a priori}, any reason to assume that such an analysis of interevent interval data through this `wrong' lens (e.g. treating the interevent times from a point process as the output from a map)   should give rise to meaningful results. However, a surprising result by Sauer~\cite{sauer1997reconstruction} demonstrates at least one scenario where this type of analysis \emph{does} give rise to meaningful results. In particular, Sauer demonstrated that when the state of a chaotic dynamical system is mapped into an interevent interval sequence via an integrate-and-fire model, a one-to-one mapping exists between the full, unobserved state of the system and an embedding of the interevent interval sequence as long as the embedding is of dimension at least twice the box counting dimension of the underlying chaotic system. Thus, it is possible to recover the true state of the entire system by considering sufficiently long interevent interval sequences.

This fact poses a problem for the analysis of interevent interval data using quantities such as Approximate Entropy or Sample Entropy, since as we have noted those approximate differential entropy rates, and the differential entropy rate of a deterministic dynamical system is negative infinity. Thus, the quantity being used is at least potentially mis-specified for the phenomenon being studied. Nevertheless, it seems unlikely that the popularity of Approximate Entropy or Sample Entropy will abate in the near future~\cite{yentes2013appropriate}, and thus it is interesting to consider how a more principled entropy rate estimator performs in the mis-specified case. Moreover, in practice the deterministic dynamical system model is almost certainly mis-specified for complex systems. As noted in~\cite{Fraser:1989ib}, there is hope that observational and dynamical noise might smooth out the infinities, thus resulting in useful estimates of entropy rates.

Consider a non-negative signal $S(t) = g(\mathbf{x}(t))$ mapping the $m$-dimensional state $\mathbf{x}(t) \in \mathbb{R}^{m}$ of a chaotic dynamical system to a scalar value. The integrate-and-fire model generates a series of discrete events based on when the integrated signal crosses a fixed threshold $\Theta$.
Setting $T_{0} = 0$, for a fixed threshold value $\Theta$, the threshold crossing events $\{ T_{i} \}$ are defined recursively as
\begin{align}
\int_{T_{i}}^{T_{i+1}} S(t) \, dt = \Theta
\end{align}
and the interevent intervals are given by the time between event $i-1$ and $i$, $\text{IEI}_{i} = T_{i} - T_{i - 1}$. 

We consider signals generated by two classic chaotic systems, the Lorenz system evolving according to
\begin{align}
\begin{split}
\dot{x} &= \sigma (y - x) \\
\dot{y} &= x( \rho - z) - y \\
\dot{z} &= x y - \beta z \\
\end{split}
\end{align}
with the canonical values $\sigma = 10, \beta = 8/3,$ and $\rho = 28$, and the R\"{o}ssler system evolving according to
\begin{align}
\begin{split}
\dot{x} &= - y - z \\
\dot{y} &= x + ay \\
\dot{z} &= b + z(x - c) \\
\end{split}
\end{align}
with the canonical values of $a = 0.1, b = 0.1,$ and $c = 14$.
For both the Lorenz and R\"{o}ssler systems, following~\cite{sauer1997reconstruction}, we take the signal to be
\begin{align}
S(t) = (x(t) + 2)^{2}
\end{align}
and fix $\Theta = 60$ and $\Theta = 125$, respectively.

\begin{figure}[!ht]
\centering
\includegraphics[width=0.75\textwidth]{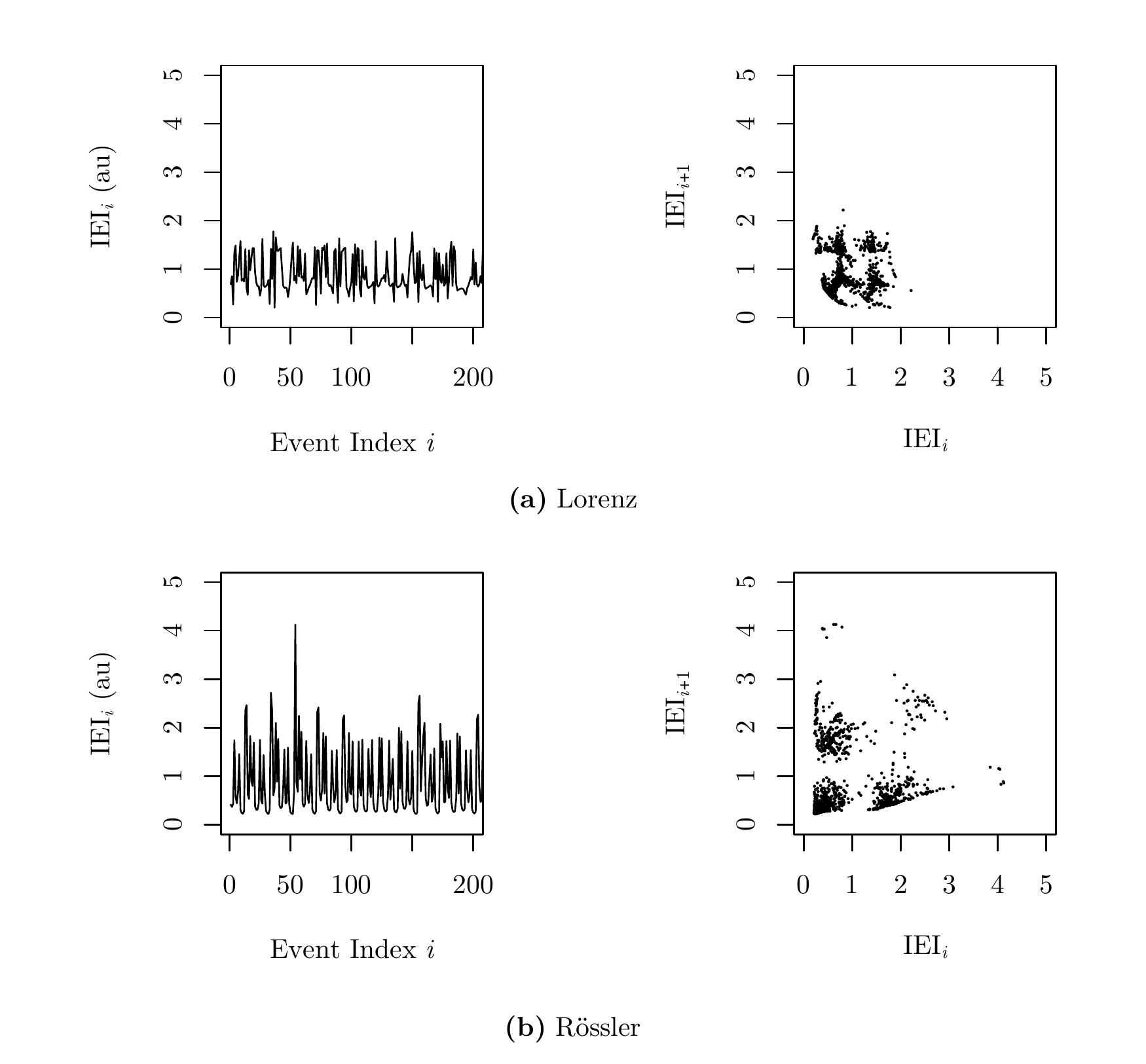}
\caption{Example interevent intervals from an integrate-and-fire model driven by the $x(t)$ states of the Lorenz (top) and R\"{o}ssler (bottom) systems. The interevent interval lengths versus the event index (left) and the lag plots of the interevent interval sequences (right) for both systems.}
\label{fig:lorenz+rossler}
\end{figure}

Figure~\ref{fig:lorenz+rossler} demonstrates example realizations of the  interevent intervals $\text{IEI}_{i} = T_{i} - T_{i - 1}$ by event index $i$ (left) as well as a lag-lag plot of consecutive interevent intervals (right) for the Lorenz (top) and R\"{o}ssler (bottom) systems. We see that the two systems give rise to very different time courses of interevent intervals, as we would expect from differing dynamics of the two systems. In particular, since both the $x$- and $y$-coordinates of the R\"{o}ssler system evolve in a nearly-linear fashion, we see that the interevent intervals are relatively regular. By comparison, the interevent intervals for the Lorenz system are much more erratic. 
Thus, we might intuitively expect for the interevent intervals from the Lorenz system to give higher specific entropy rates than the interevent intervals from the R\"ossler system.

\begin{figure}[!ht]
\centering
\includegraphics[width=0.6\textwidth]{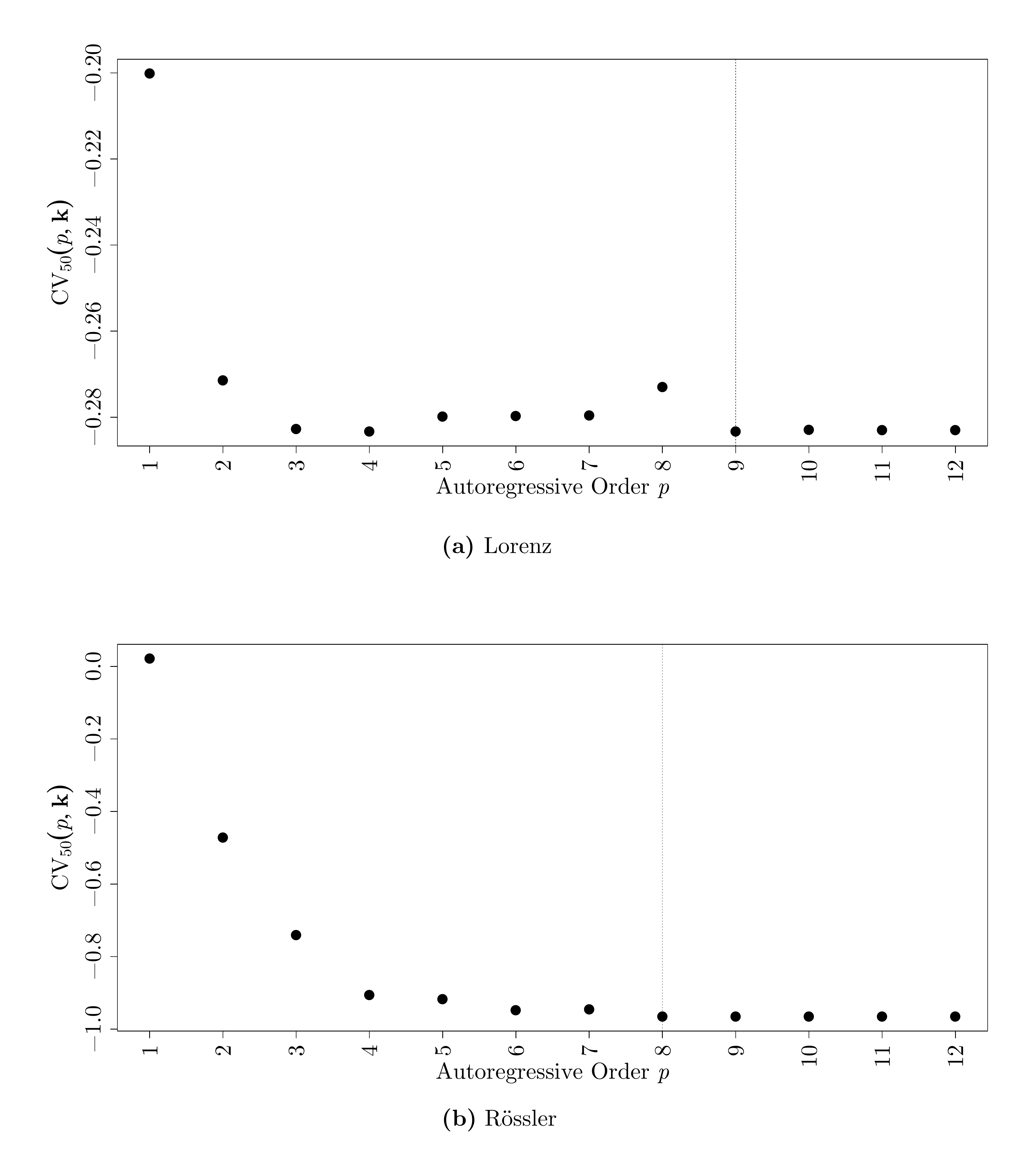}
\caption{The 50-block cross-validated log-likelihoods (\ref{eqn:cv-function}) for the Lorenz (top) and R\"{o}ssler (bottom) interevent interval sequences as a function of the autoregressive order $p$. The vertical lines mark the minimum 50-block cross-validated log-likelihoods which occur at $p = 9$ and $p = 8$, respectively.}
\label{fig:lorenz+rossler-choose-p}
\end{figure}

Next we turn to estimating the specific entropy rate for each of these systems. For each system, we generated interevent interval sequences of length $T = 1000$. We then chose the model order $p$ and bandwidths $(k_{1}, \ldots, k_{p+1})$ as described in Section~\ref{sec:model-selection}. The 50-block cross-validated log-likelihood (\ref{eqn:cv-function}) as a function of $p$ is shown in Figure~\ref{fig:lorenz+rossler-choose-p}. Based on the embedology~\cite{Sauer:1991kv} result from~\cite{sauer1997reconstruction}, an embedding of at least twice the box counting dimension of the underlying attractor is required. Both the Lorenz and R\"ossler attractors have box counting dimensions between 2 and 3, thus we expect that a value of $p$ around 6 should be sufficient for the predictive density. We see that the 50-block cross-validated log-likelihood chooses $p = 9$ and $p = 8$ for the Lorenz and R\"ossler systems.

As mentioned in Sections~\ref{sec:kde} and~\ref{sec:model-selection}, using cross-validation to choose the bandwidths of the conditional kernel density estimator introduces a form of feature selection into the conditional density estimation process: lags that are not relevant, as measured by the cross-validation score, are smoothed out by setting their associated bandwidths to infinity (in practice, to a large value). We demonstrate this phenomenon now for the bandwidths estimated for the interevent intervals derived from the Lorenz and R\"ossler systems. For a fixed maximal lag $p$, Table~\ref{table:bws-lorenz+rossler} shows the bandwidths estimated for the Lorenz (top) and R\"ossler (bottom) systems. The first row indicates the bandwidths chosen by cross-validation for the future $k_{0}$ and past $k_{-1}$ when we include only a single lag, the second row indicates the bandwidths chosen for the future $k_{0}$ and past $(k_{-1}, k_{-2})$ when we include two lags, etc. A horizontal dash (---) indicates that cross-validation has set the bandwidth associated with that lag to a value of 5 or greater, which is large with respect to the scale of the dynamics, thus in effect ignoring the lag in the estimation of the predictive density. Note that these bandwidths are for Gaussian kernels, and thus are not immediately at the scale of the data. A transformation from the Gaussian scale to the uniform scale could be performed using the concept of canonical kernels~\cite{marron1988canonical}.  Comparing Table~\ref{table:bws-lorenz+rossler} to Figure~\ref{fig:lorenz+rossler-choose-p}, we see that for the interevent intervals generated by the Lorenz system, intervals 4 through 7 can be ignored. This agrees with the sharp drop in Figure~\ref{fig:lorenz+rossler-choose-p} at $p = 3$. Then, the intervals 8 and 9 are included, but no others, thus giving the minimum at $p = 9$. A similar result holds for the bandwidths for the R\"ossler-governed interevent intervals, where the bandwidths stabilize at $p = 8$, which also corresponds to the minima in the 50-block cross-validated log-likelihood. Beyond this automatic selection of relevant lags, we see that the \emph{magnitudes} of the bandwidths are very different amongst the $\mathbf{k} = (k_{0}, k_{-1}, \ldots, k_{-p})$: as one might expect, the bandwidths for the near past are smaller than the bandwidths for the distant past, \emph{i.e.} we should pay more attention to the recent past for prediction. Compare this inherent dynamic range in the bandwidths across lags to the fixed bandwidths across lags used in other statistics such as Approximate Entropy, Sample Entropy, and Multiscale Entropy. If viewed as estimators of different differential entropy rates, these estimators would be severely biased by the fixed bandwidths.

\begin{table}
	\caption{The optimal bandwidths $\mathbf{k} = (k_{0}, k_{-1}, \ldots, k_{-p})$ chosen using~(\ref{eqn:cv-function}) with $p$ fixed from 1 to 12 for the interevent intervals derived from the Lorenz (top) and R\"ossler (bottom) systems. A horizontal dash (---) indicates that cross-validation set the bandwidth associated with that lag to a value of 5 or greater, in effect ignoring the lag in the estimation of the predictive density. The bold rows correspond to bandwidths selected for the minimal values of $p$ as shown in Figure~\ref{fig:lorenz+rossler-choose-p}.}
	\tiny
\begin{subtable}[t]{6 in}
\begin{tabular}{c || c | c c c c c c c c c c c c }
$p$ & $k_{0}$ & $k_{-1}$ & $k_{-2}$ & $k_{-3}$ & $k_{-4}$ & $k_{-5}$ & $k_{-6}$ & $k_{-7}$ & $k_{-8}$ & $k_{-9}$ & $k_{-10}$ & $k_{-11}$ & $k_{-12}$ \\ \hline
1 & 0.048 & 0.035 &  &  &  &  &  &  &  &  &  &  &  \\
2 & 0.059 & 0.039 & 0.055 &  &  &  &  &  &  &  &  &  &  \\
3 & 0.059 & 0.039 & 0.051 & 0.559 &  &  &  &  &  &  &  &  &  \\
4 & 0.059 & 0.039 & 0.051 & 0.558 & --- &  &  &  &  &  &  &  &  \\
5 & 0.059 & 0.039 & 0.051 & 0.563 & --- & --- &  &  &  &  &  &  &  \\
6 & 0.059 & 0.039 & 0.051 & 0.564 & --- & --- & --- &  &  &  &  &  &  \\
7 & 0.059 & 0.039 & 0.051 & 0.576 & --- & --- & --- & --- &  &  &  &  &  \\
8 & 0.070 & 0.050 & 0.057 & 0.450 & 0.541 & 0.625 & --- & --- & 0.674 &  &  &  &  \\
\textbf{9} & \textbf{0.059} & \textbf{0.039} & \textbf{0.052} & \textbf{0.570} & \textbf{---} & \textbf{---} & \textbf{---} & \textbf{---} & \textbf{1.263} & \textbf{0.826} &  &  &  \\
10 & 0.059 & 0.039 & 0.052 & 0.573 & --- & --- & --- & --- & 1.194 & 0.816 & --- &  &  \\
11 & 0.059 & 0.039 & 0.052 & 0.571 & --- & --- & --- & --- & 1.188 & 0.819 & --- & --- &  \\
12 & 0.059 & 0.039 & 0.052 & 0.574 & --- & --- & --- & --- & 1.184 & 0.816 & --- & --- & --- \\
\end{tabular}
\centering \caption{Lorenz}
\label{table:bws-lorenz+rossler-lorenz}
\end{subtable}
\\ \\ \\
\begin{subtable}[t]{6 in}
\begin{tabular}{c || c | c c c c c c c c c c c c }
$p$ & $k_{0}$ & $k_{-1}$ & $k_{-2}$ & $k_{-3}$ & $k_{-4}$ & $k_{-5}$ & $k_{-6}$ & $k_{-7}$ & $k_{-8}$ & $k_{-9}$ & $k_{-10}$ & $k_{-11}$ & $k_{-12}$ \\ \hline
1 & 0.047 & 0.087 &  &  &  &  &  &  &  &  &  &  &  \\
2 & 0.062 & 0.054 & 0.052 &  &  &  &  &  &  &  &  &  &  \\
3 & 0.064 & 0.049 & 0.044 & 0.058 &  &  &  &  &  &  &  &  &  \\
4 & 0.065 & 0.048 & 0.046 & 0.072 & 0.078 &  &  &  &  &  &  &  &  \\
5 & 0.065 & 0.049 & 0.047 & 0.073 & 0.087 & 0.575 &  &  &  &  &  &  &  \\
6 & 0.065 & 0.053 & 0.051 & 0.082 & 0.089 & 0.751 & 0.185 &  &  &  &  &  &  \\
7 & 0.064 & 0.052 & 0.051 & 0.088 & 0.086 & 0.787 & 0.359 & 0.732 &  &  &  &  &  \\
\textbf{8} & \textbf{0.065} & \textbf{0.053} & \textbf{0.055} & \textbf{0.086} & \textbf{0.100} & \textbf{---} & \textbf{0.360} & \textbf{0.820} & \textbf{0.553} &  &  &  &  \\
9 & 0.064 & 0.054 & 0.055 & 0.086 & 0.100 & --- & 0.366 & 0.805 & 0.613 & --- &  &  &  \\
10 & 0.065 & 0.053 & 0.054 & 0.085 & 0.100 & --- & 0.359 & 0.810 & 0.573 & --- & --- &  &  \\
11 & 0.064 & 0.054 & 0.055 & 0.087 & 0.099 & --- & 0.369 & 0.812 & 0.592 & --- & --- & --- &  \\
12 & 0.065 & 0.054 & 0.054 & 0.086 & 0.101 & --- & 0.366 & 0.808 & 0.580 & --- & --- & --- & --- \\
\end{tabular}
\centering  \caption{R\"ossler.}
\label{table:bws-lorenz+rossler-rossler}
\end{subtable}
\label{table:bws-lorenz+rossler}
\end{table}

\begin{figure}[!ht]
	\centering
	\includegraphics[width=1\textwidth]{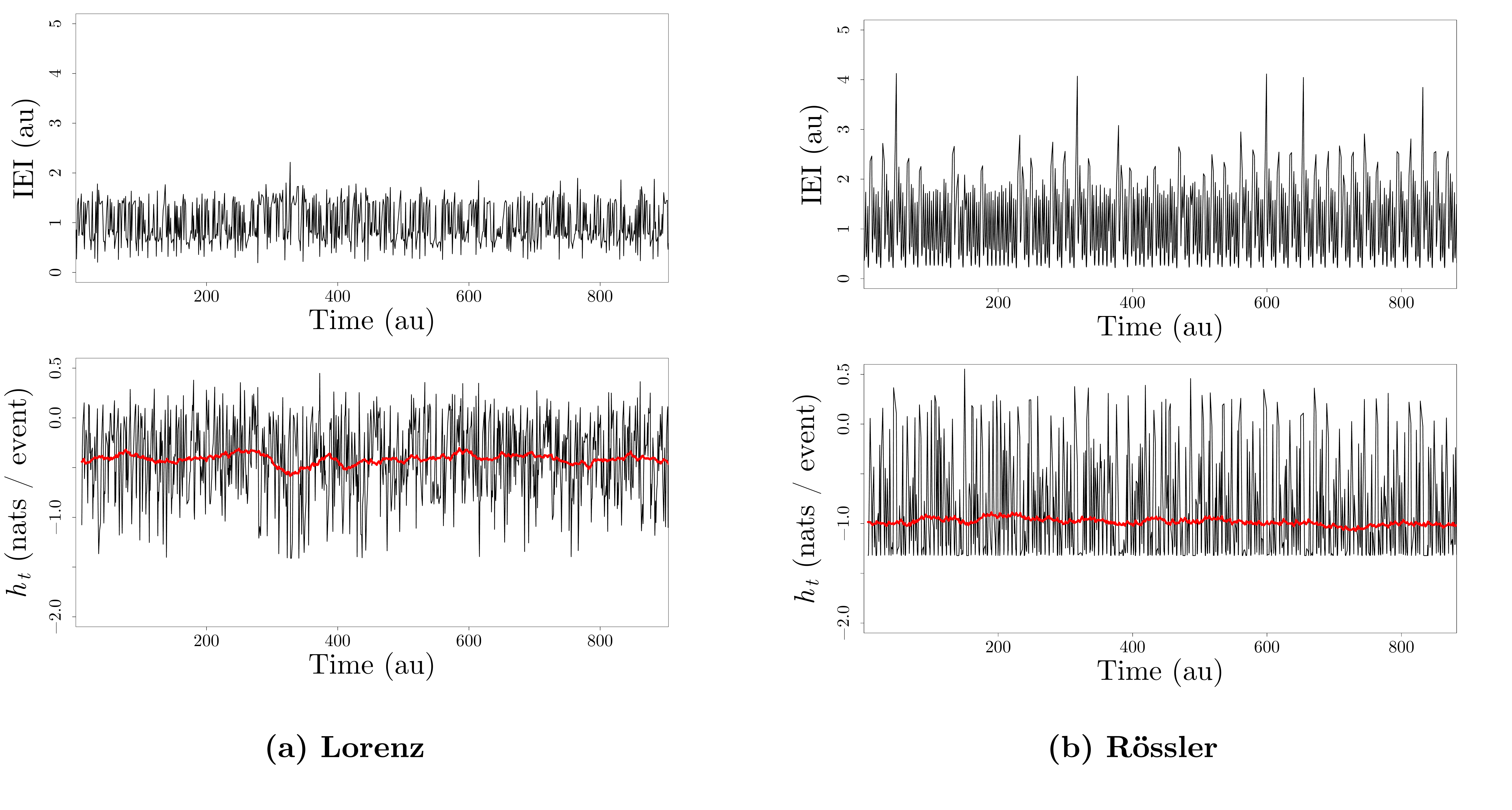}
	\caption{The interevent interval sequence (top) and specific entropy rate (bottom) for the Lorenz (left) and R\"ossler (right) systems. Note that both the interevent intervals and specific entropy rates are plotted as a function of the event times rather than the event index. The solid red line indicates a time-windowed average of the specific entropy rate with a uniform kernel with window length of 60 au.}
	\label{Fig-lorenz+rossler-singletons}
\end{figure}

Now consider the specific entropy rate of two interevent interval sequences as a function of time, shown in Figure~\ref{Fig-lorenz+rossler-singletons}. Note that both the interevent intervals and specific entropy rates are shown as a function of the \emph{time} rather than the \emph{event} index. That is, for each interevent interval sequence, we show $(T_{i}, \text{IEI}_{i})$ and $(T_{i}, h_{i})$. The estimate of the time-averaged specific entropy rate~(\ref{eqn:final-er-estimator}) for the Lorenz and R\"ossler interevent interval sequences are $-0.41$ nats / event and $-1.0$ nat / event, respectively. In addition, we also show a moving windowed average of the specific entropy rate using a uniform kernel of width 60 au in red in the bottom panel of  Figure~\ref{Fig-lorenz+rossler-singletons}. This can be thought of as a local (in time) version of~(\ref{eqn:final-er-estimator}), and allows us to determine if there are periods of time when the interevent intervals are more, or less, predictable. For example, we see a drop in the specific entropy rate for the Lorenz interevent intervals around 300 au, which corresponds to a run of relatively long and regular interevent intervals.

We see from both~(\ref{eqn:final-er-estimator}) and its time-local version that the interbeat interval sequence derived from the R\"ossler system are more predictable, which matches our intuition as outlined above based on the near-linear dynamics of the $x$-coordinate of the R\"ossler system. The thresholds $\Theta$ were chosen such that each system has approximately equal mean interevent interval length: $0.90$ au and $0.88$ au for the Lorenz and R\"ossler systems, respectively. However, the pointwise standard deviations of the two interevent interval sequences \emph{are} different: $0.39$ au and $0.73$ au for the Lorenz and R\"ossler systems, respectively. Recall that, unlike discrete entropy, differential entropy is \emph{not} scale invariant. This motivates determining a scale invariant analog of the specific entropy rate that teases apart inherent unpredictability from the natural scale of the system. We will consider this point in the discussion section.

\begin{figure}[!ht]
	\centering
	\includegraphics[width=0.5\textwidth]{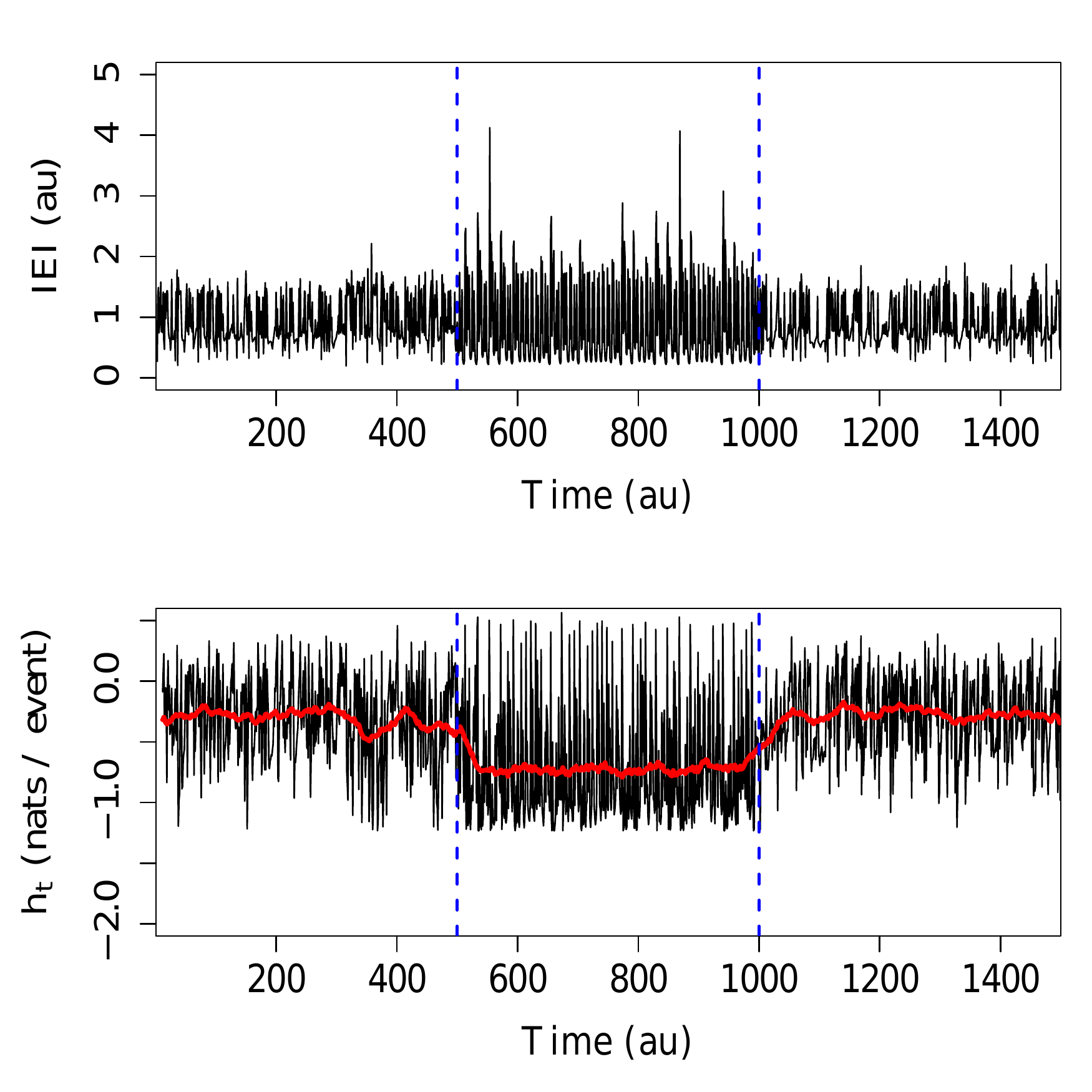}
	\caption{The interevent interval sequence (top) and specific entropy rate (bottom) for the concatenation of Lorenz, R\"ossler, and Lorenz interevent intervals. The dashed blue lines indicate the transitions from one system to the other. Compare to Figure~\ref{Fig-lorenz+rossler-singletons}, where the specific entropy rates were estimated individually for each system.}
	\label{Fig-lorenz+rossler-concatenated}
\end{figure}

\begin{figure}[!ht]
	\centering
	\includegraphics[width=0.6\textwidth]{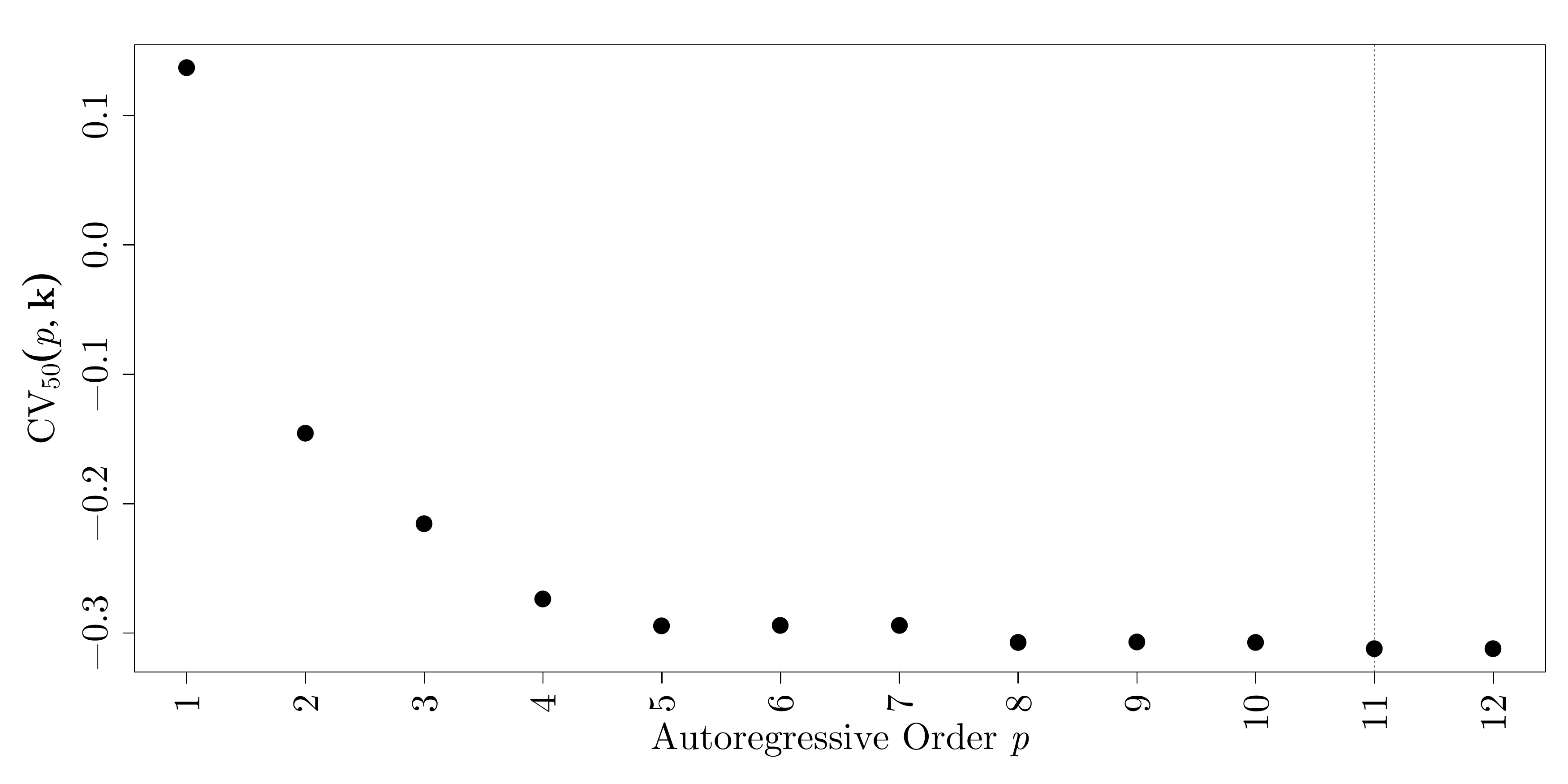}
	\caption{The 50-block cross-validated log-likelihood (\ref{eqn:cv-function}) for the concatenation of the Lorenz, R\"ossler, and Lorenz interevent interval sequences as a function of the autoregressive order $p$. The vertical line marks the minimum log-likelihood which occurs at $p = 11$.}
	\label{fig:lorenz+rossler+lorenz-choose-p}
\end{figure}

As a final example, we consider estimation of the specific entropy rate where the interevent interval sequence transitions from being generated by the Lorenz system to being generated by the R\"ossler system and back again. In this case, the interevent interval sequence is clearly non-stationary. However, conditional stationarity is only violated locally in time around the transitions. To generate this time series, we concatenate 500 interevent intervals each from the Lorenz, Rossler, and Lorenz systems, and thus $T = 1500$. This sequence is shown in the top panel of Figure~\ref{Fig-lorenz+rossler-concatenated}. We estimate the autoregressive order $p$ over the entire time series using (\ref{eqn:cv-function}). The 50-block cross-validated log-likelihood as a function of $p$ is shown in Figure~\ref{fig:lorenz+rossler+lorenz-choose-p}. The minima occurs at $p = 11$. Note that this is a higher order than chosen for either the Lorenz ($p = 9$) or R\"ossler ($p = 8$) systems when estimated in isolation. We see that additional information about the past is required when we need to distinguish between the two systems. Finally, Table~\ref{table:bws-lorenz+rossler+lorenz} demonstrates the bandwidths chosen by cross-validation as a function of the maximal lag $p$. Again, we see that cross-validation provides both model selection and adaptive smoothing.

\begin{table}
\caption{The optimal bandwidths $\mathbf{k} = (k_{0}, k_{-1}, \ldots, k_{-p})$ chosen using~(\ref{eqn:cv-function}) with $p$ fixed from 1 to 12 for the interevent intervals derived from the concatenation of the Lorenz, then R\"ossler, then Lorenz systems. A horizontal dash (---) indicates that cross-validation set the bandwidth associated with that lag to a value of 5 or greater, in effect ignoring the lag in the estimation of the predictive density. The bold row correspond to bandwidths selected for the minimal value of $p$ as shown in Figure~\ref{fig:lorenz+rossler+lorenz-choose-p}.}
\tiny
\begin{tabular}{c || c | c c c c c c c c c c c c }
$p$ & $k_{0}$ & $k_{-1}$ & $k_{-2}$ & $k_{-3}$ & $k_{-4}$ & $k_{-5}$ & $k_{-6}$ & $k_{-7}$ & $k_{-8}$ & $k_{-9}$ & $k_{-10}$ & $k_{-11}$ & $k_{-12}$ \\ \hline
1 & 0.048 & 0.063 &  &  &  &  &  &  &  &  &  &  &  \\
2 & 0.064 & 0.046 & 0.059 &  &  &  &  &  &  &  &  &  &  \\
3 & 0.074 & 0.046 & 0.047 & 0.370 &  &  &  &  &  &  &  &  &  \\
4 & 0.071 & 0.049 & 0.051 & 0.417 & 0.459 &  &  &  &  &  &  &  &  \\
5 & 0.070 & 0.047 & 0.058 & 0.431 & 0.512 & 0.650 &  &  &  &  &  &  &  \\
6 & 0.070 & 0.047 & 0.058 & 0.431 & 0.513 & 0.649 & --- &  &  &  &  &  &  \\
7 & 0.070 & 0.047 & 0.058 & 0.432 & 0.513 & 0.646 & --- & --- &  &  &  &  &  \\
8 & 0.070 & 0.050 & 0.057 & 0.450 & 0.541 & 0.625 & --- & --- & 0.674 &  &  &  &  \\
9 & 0.070 & 0.051 & 0.059 & 0.455 & 0.531 & 0.661 & --- & --- & 0.710 & --- &  &  &  \\
10 & 0.070 & 0.050 & 0.057 & 0.454 & 0.542 & 0.620 & --- & --- & 0.666 & --- & --- &  &  \\
\textbf{11} & \textbf{0.071} & \textbf{0.051} & \textbf{0.058} & \textbf{0.470} & \textbf{0.548} & \textbf{0.632} & \textbf{---} & \textbf{---} & \textbf{0.622} & \textbf{---} & \textbf{---} & \textbf{0.985} &  \\
12 & 0.071 & 0.051 & 0.057 & 0.471 & 0.548 & 0.634 & --- & --- & 0.628 & --- & --- & 0.997 & --- \\
\end{tabular}
\label{table:bws-lorenz+rossler+lorenz}
\end{table}

The bottom panel of Figure~\ref{Fig-lorenz+rossler-concatenated} shows the specific entropy rate as a function of time for the concatenated system. As before, the black line is the specific entropy rate, and the red line is a moving windowed average of the specific entropy rate. Again we see that the specific entropy rate drops as the system transitions from the Lorenz interevent intervals to the R\"ossler interevent intervals and then increases after the transition back to the Lorenz interevent intervals. There is, however, a slight penalty to estimating the specific entropy rate for the concatenated interevent interval sequences all at once. During the Lorenz-governed interevent interval sequence, the time-averaged specific entropy rates are $-0.30$ nats / event and $-0.28$ nats / event, compared to $-0.41$ nats / event when estimated in isolation. Similarly, the time-averaged specific entropy rate for the R\"ossler-governed interevent interval sequence is $-0.72$ nats / event compared to $-1.0$ nats / event when estimated in isolation. In both cases, we see that the specific entropy rates have increased.  This is largely due to the fact that the optimal bandwidths $k_{1}, \ldots, k_{p+1}$ when estimating the predictive density for either system in isolation are \emph{not} optimal for estimating the concatenation of the two systems. This will lead to larger bandwidths overall, and thus higher specific entropy rates. For this system, the difference in the dynamics is very large and the transition point relatively obvious, and thus a better approach might be to estimate the predictive densities separately for each segment. However, in those cases where such transitions are non-obvious or where manual transition detection is not desirable, we see that estimating the predictive density all at once still leads to discrimination between high and low specific entropy rates.

\begin{figure}[!ht]
	\centering
	\includegraphics[width=1\textwidth]{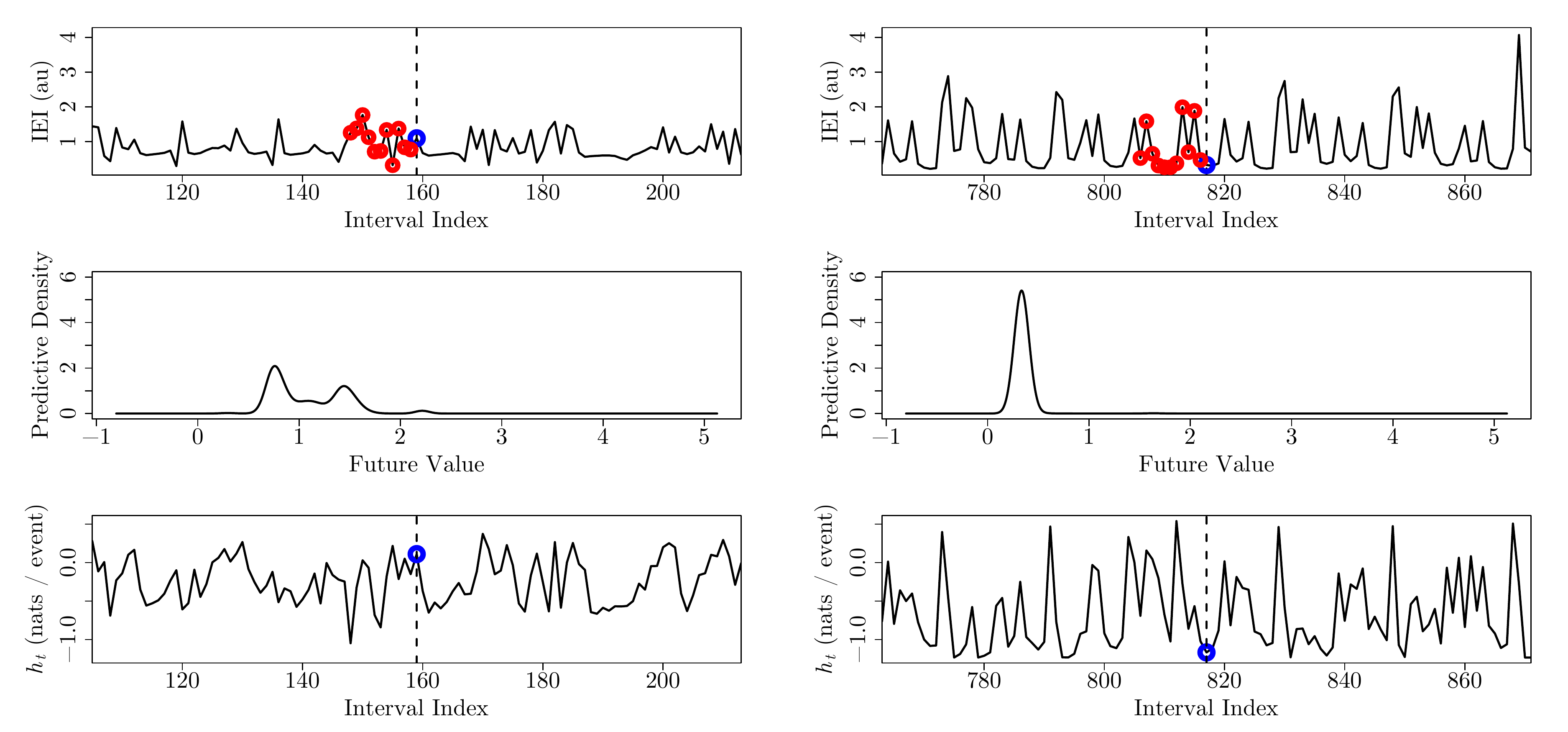}
	\caption{A demonstration of (top) the interevent interval sequence, (middle) the estimated predictive density $\hat{f}\left(\text{IEI}_{i} \mid \text{IEI}_{i-11}^{i-1}\right)$, and (bottom) the specific entropy rate for the concatenated Lorenz, R\"ossler, Lorenz system during the Lorenz (left) and R\"ossler (right) portions of the sequence. In the top panels, the dashed vertical bar indicates the event index $i$, the red circles correspond to the specific past $\text{IEI}_{i-11}^{i-1}$, and the blue circles correspond to the future value $\text{IEI}_{i}$.}
	\label{Fig-schematic-lorenz+rossler+lorenz}
\end{figure}

Figure~\ref{Fig-schematic-lorenz+rossler+lorenz} demonstrates the interevent interval sequence (top), predictive density (middle), and specific entropy rate (bottom) for interevent interval sequence for two time instants during the Lorenz (left) and R\"ossler (right) portions. The time instant during the portion governed by the Lorenz system has a higher specific entropy rate, as we would expect given the multi-modal nature of the estimated predictive density in the middle panel. In contrast, the time instant during the portion governed by the R\"ossler system has a lower specific entropy rate, as we would expect from the uni-modal and narrow estimated predictive density. However, we see that in both cases, the specific entropy rate can vary widely depending on the state of the system. For example, during periods around the long interevent intervals, the interevent intervals generated by the R\"ossler system can have higher specific entropy rates than those governed by the Lorenz system (the peaks in the specific entropy rate).

\subsection{Specific Entropy Rate from a Tilt Table Experiment}

\label{sec:hrv}

As a last example, we consider the specific entropy rates of interbeat interval sequences from subjects participating in a tilt table experiment. It is well known by anyone with a heart that the \emph{rate} of their pulse, the average number of beats within a specified window of time, can vary widely based on environmental, physiological, and psychological factors.  However, it was not until the 20th century that researchers came to realize that \emph{beat-to-beat} variations in heart rate convey information about the health of individuals. The study of beat-to-beat variations in heart rate is typically referred to under the umbrella term of heart rate variability. See ~\cite{RajendraAcharya:2006kq,Berntson:2007et,Billman:2011gva} for a historical perspective on heart rate variability. The nonlinear dynamics community has contributed a large number of methods for the analysis of interbeat intervals. See~\cite{Voss:2009km} for an extensive historical and methodological review.

In what follows, we use the term interbeat interval (IBI) to refer to the times between the R components of adjacent QRS complexes associated with heartbeats. Common statistics computed from heart rate variability data include the mean interbeat interval and the standard deviation of the interbeat intervals. In addition, it is common to interpolate the interbeat interval sequence to obtain an equi-spaced sequence for spectral analysis~\cite{Deboer:1984OfXaGbAE}, from which the power of high frequency and low frequency components, and their ratio, are commonly reported. It is also very common to compute Approximate and / or Sample Entropies of interbeat interval sequences. Any, and sometimes all, of these statistics are referred to as heart rate variability (HRV), and thus we will refrain from using that term. Many of these quantities can be computed by off-the-shelf software tailored for heart rate variability analysis such as Kubios~\cite{tarvainen2014kubios}, though we recommend caution when using such software since many of the parameters involved in both pre-processing of the data and its analysis are set in an \emph{ad hoc} fashion.

As before, our approach to analyzing an interbeat interval sequence is to view it as the realization of some conditionally stationary stochastic dynamical system. This perspective naturally handles the fact that heart beats occur as a point process in time, as we saw in the previous section. Thus, we can compute the specific entropy rate associated with the time until the next heart beat, conditional on the most recent interbeat intervals. That is, if we denote the time between the $(i - 1)^\text{th}$ and $i^\text{th}$ heart beat by $\text{IBI}_{i}$, we consider the specific entropy rate as $h\left[\text{IBI}_{i} \mid \text{IBI}_{i - p}^{i - 1}\right].$

We will investigate the specific entropy rate from the interbeat interval sequences of five subjects participating in a tilt table experiment. The population consisted of two males and three females between the ages of 27 and 44. In the experiment, the subject initially positioned him/herself in a prone position on the table and was secured to the table. The subject was then kept in the supine position for 5 minutes, then tilted upright for 5 minutes, and finally was returned to a supine position for 5 minutes. An ECG was continuously recorded throughout the experiment. The interbeat intervals were extracted using the AF1 algorithm from~\cite{Friesen:1990ce}.

\begin{figure}[!ht]
	\centering
	\includegraphics[width=1\textwidth]{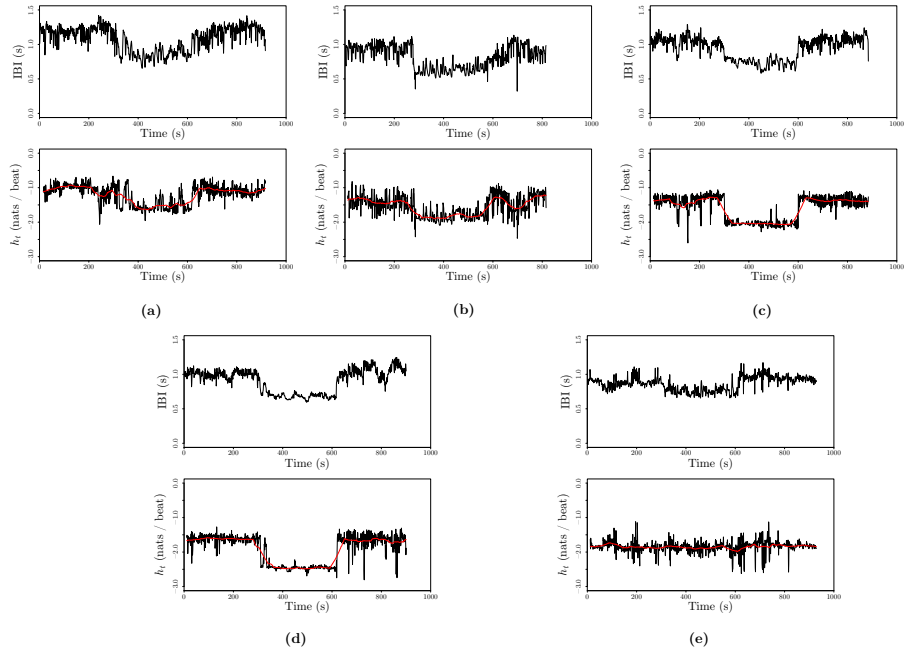}
	\caption{The interbeat interval sequences (top) and specific entropy rates (bottom) for each of the five subjects (a-e) in the tilt table experiment. The solid red line indicates a time-windowed average of the specific entropy rate with a uniform kernel with window length of 60 s.}
	\label{Fig-tilt-table-ers}
\end{figure}

Specific entropy rates were computed for each subject using model orders $p$ and bandwidths $(k_{1}, \ldots, k_{p+1})$ chosen as described in Section~\ref{sec:model-selection}. The interbeat interval sequences (top) and specific entropy rates (bottom) for each subject are shown in Figure~\ref{Fig-tilt-table-ers}. For each subject, we see the expected decrease in interbeat interval length (increase in heart rate) as they move from a supine to upright position. However, for subjects (a-d), this change in mean interbeat interval length is also associated with a change in the overall dynamics of the interbeat interval sequence, which results in a drop in the specific entropy rate during the upright time period. With the return to supine position, the interbeat interval lengths again increase (the heart rate decreases), and the specific entropy rates of subjects (a-d) return to the same level as the start of the experiment.

Clearly, with only five subjects and a single session from each subject, we cannot say much about either typical or atypical evolution of specific entropy rates in a tilt table experiment. However, it is interesting to note that subject (e), the only outlier in terms of the evolution of their specific entropy rate over time, is also the only subject with a traumatic brain injury in their past. Head trauma has been associated with changes in both spectral and information theoretic properties of interbeat interval sequences at rest~\cite{Su:2005je,Papaioannou:2008hr}. Our results corroborate these findings, and suggest that additional studies that include a physiological stressor, such as the tilt table, may be even more disclosing.






\section{Discussion and Future Directions}

\label{sec:discussion}

An important consideration for any estimator relates to how it behaves under error or in the presence of noise.  Care must be taken with respect to how one defines error, however. For example, does error refer to observational noise, model uncertainty / misspecification, or unobserved factors~\cite{crutchfield2003regularities}? We have not considered the impact of observational noise, for example, because the measurements we have considered, namely interevent and interbeat intervals, can be treated as relatively noise free. However, if observational noise \emph{is} a major concern, then the estimation of specific entropy rate must be carefully applied in this context, since direct estimation from the observed signal will combine dynamical and observational uncertainties. Possible solutions include the errors-in-variables model for density estimation~\cite{bosq2012nonparametric} or more general nonlinear filtering approaches~\cite{tanizaki1996nonlinear}.

We have considered only \emph{fixed} bandwidths for the conditional kernel density estimator in estimating the specific entropy rate: regardless of the past and future states of the system, we use the same bandwidths in estimating the predictive density. In Section~\ref{sec:chaotic-ii}, we saw a scenario where this estimation strategy may be problematic: the typical scale of the interevent intervals differed between the Lorenz-governed and R\"ossler-governed periods, and this led to suboptimal bandwidths. Alternative variable bandwidth density estimation schemes allow the bandwidths to vary with either the data used in estimation of the density or the point of evaluation~\cite{Terrell:1992}. For example, the estimator for the differential entropy of a random vector developed in~\cite{kozachenko1987sample} based on $k$-nearest neighbor statistics is equivalent to a plug-in estimator of the differential entropy using a kernel density estimator with a bandwidth that varies with the point of evaluation, in this case the distance to the $k^{\text{th}}$ nearest neighbor of the evaluation point, along with an additional bias correction term. Many other estimators, such as the popular Kraskov-St\"ogbauer-Grassberger mutual information estimator~\cite{Kraskov:2004gr}, fall into this category. Future work will explore the tradeoff between the resolution gained by variable bandwidth estimators of specific entropy rate and the statistical and computational burden imposed. One recent approach along these lines used a variable bandwidth kernel density estimator to estimate the transfer entropy for various simulated systems~\cite{Zuo:ef}.

Another potential issue with scaling, as we again saw in Section~\ref{sec:chaotic-ii}, is that differential entropy, and thus differential entropy rate, is not invariant to scaling. For example, changing the units used to measure the system under consideration will result in a linear shift to the differential entropy. Depending on the application at hand, this may or may not be problematic. If comparing the entropy rate of multiple time series, all with the same units, then the lack of invariance to scale washes out. However, if one is analyzing a single time series that has large variations in its characteristic scale over time, then the dependence on scaling may be problematic. One potential alternative is to normalize the differential entropy rate using the typical scale of the system at any given instant. A good candidate for this is the \emph{negentropy}~\cite{Comon:1994kr} of a random variable, which normalizes the differential entropy by the differential entropy associated with a Gaussian density with the same variance. The negentropy, unlike the differential entropy, is invariant to affine transformations of a random variable. Thus, we might define a \emph{specific negentropy rate} by normalizing the specific entropy rate by an instantaneous measure of the variance. This is analogous to the \emph{redundancy}~\cite{crutchfield2003regularities} of discrete-state stochastic processes, which normalizes the entropy rate of a stochastic process by the entropy rate of a uniformly distributed process with the same alphabet.

Any method that utilizes either Approximate or Sample Entropies could be modified to use our specific entropy rate estimator. For example, the multiscale entropy~\cite{Costa:2002em}, which is defined as the Sample Entropy of a time series at varying levels of aggregation, could easily be modified by direct substitution with the specific entropy rate. This would allow for not only an analysis of the unpredictability across \emph{scales}, but also across \emph{time}. Similarly, the point process model of interbeat interval sequences introduced in~\cite{Barbieri:2004ci,ZheChen:2010hk,Valenza:2013ih} is a particular parametric form for the stochastic dynamical system (\ref{eqn:transition-sds}). In a sequel~\cite{Valenza:2014ie}, the authors propose using the filtered state from this model to estimate what they call the inhomogeneous point-process entropy. They estimate this quantity using either the Approximate or Sample Entropies, and thus based on the analysis from~\cite{lake2006renyi}, we see that their estimator is for the unscaled Shannon or R\'enyi entropy rate of the filtered state. Thus, the specific entropy rate could be used on the filtered state.

Our approach to \emph{specific} entropy rate estimation via conditional kernel density estimation can also be extended to any of the various other information theoretic measures gaining popularity including transfer entropy~\cite{Schreiber:2000jx,kaiser2002information}, causation entropy~\cite{Sun:2014ee}, and co-/multi-informations~\cite{bell2003co}. Many of these quantities would benefit from a data-driven approach to bandwidth selection, in addition to the automatic dimension reduction such approaches induce. However, we also note that with each additional probabilistic conditioning required by these measures, we increase both the statistical and computational burden for constructing the appropriate estimator. For example, the convergence rate of kernel density-based estimators for many information theoretic quantities scale exponentially in the reciprocal of the dimension of the random vector~\cite{kandasamy2014influence}, while their time complexities scale exponentially in the dimension of the random vector~\cite{Singh:2016vk}.
\section{Conclusions}

Via a decomposition of the entropy rate of a discrete-time, continuous-valued stochastic dynamical system, we have proposed a measure of state-specific uncertainty: the specific entropy rate. We have shown how to estimate the specific entropy rate from finite data using kernel density estimators, and provided a data-driven method for choosing the free parameters in the kernel density estimation. Given the immense popularity of heuristic approaches to entropy rate estimation such as Approximate Entropy and Sample Entropy, it is our hope that a more principled approach to entropy rate estimation will be found useful by the larger research community.

All of the software used in this paper was developed in R via extensions to the \texttt{np} library for kernel density estimation. We plan to make this implementation available online. In an effort to match the naming convention applied to Approximate Entropy (ApEn) and Sample Entropy (SampEn), we call our R implementation \texttt{spenra} for sp(ecific) en(tropy) ra(te). The R implementation of \texttt{spenra} will be hosted at \url{http://github.com/ddarmon/spenra}.

\section{Acknowledgements}

The author thanks C. Wang, D. Keyser, C. Cellucci, and P. Rapp for valuable discussions, and D. Nathan for providing the data from the tilt table experiment.

\section{Appendix -- Relationship Between the Kernel Density Estimator for Differential Entropy Rate and Approximate Entropy}

\label{sec:ApEn}

In this appendix, we make the connection first noted in~\cite{lake2006renyi} between the kernel density estimator for differential entropy rate and Approximate Entropy, emphasizing the implicit assumptions on the kernel, bandwidths, etc., that result from the default parameters used by most Approximate Entropy-based analyses. However, we also note that~\cite{pincus1991approximate} did not motivate Approximate Entropy as a kernel density-based estimator of entropy rate, but rather as a family of statistics for comparing two time series. This explains, for example, the inclusion of both self-matching and sample size independent bandwidths, which would lead to estimation bias from the perspective of kernel density estimation.

We begin by recalling the standard formulation of Approximate Entropy from~\cite{pincus1991approximate}. Consider a time series $\left\{X_{t}\right\}_{t = 1}^{T}$. For an  embedding dimension $p$, we form the embedding vectors $\left\{\mathbf{X}^{(p)}_{t}\right\}_{t=1}^{T-p+1}$ where $\mathbf{X}^{(p)}_{t} = (X_{t}, X_{t+1}, \ldots, X_{t + p - 1})$. For each vector $\mathbf{X}^{(p)}_{t}$, we compute the number of other vectors (including the vector indexed by $t$) that are within a tolerance $r$ of $\mathbf{X}^{(p)}_{t}$ under the infinity norm,
\begin{align}
	C_{t}^{(p)}(r) = \frac{\# \left\{ \mathbf{X}^{(p)}_{t'} : \left|\left|\mathbf{X}^{(p)}_{t} - \mathbf{X}^{(p)}_{t'}\right|\right|_{\infty} \leq r\right\}}{T - p + 1}, \label{Eqn:cond-prob}
\end{align}
where we recall that the infinity norm $||\cdot||_{\infty}$ of a vector $\mathbf{u} = (u_{1}, \ldots, u_{p})$ is given by
\begin{align}
||\mathbf{u}||_{\infty} = \max_{i} |u_{i}|.
\end{align}
Finally, we compute the average logarithm of (\ref{Eqn:cond-prob}) across all of the vectors, giving
\begin{align}
	\Phi^{(p)}(r) = \frac{1}{T - p + 1}\sum_{t = 1}^{T-p+1} \log C_{t}^{(p)}. \label{Eqn:def-Phi}
\end{align}
For fixed $p, r,$ and $T$, the Approximate Entropy is defined as
\begin{align}
	\text{ApEn}(p, r, T) = \Phi^{(p)}(r) - \Phi^{(p + 1)}(r). \label{Eqn:ApEn}
\end{align}

We next show that (\ref{Eqn:ApEn}) is almost equivalent to a plug-in entropy rate estimator based on kernel density estimation. We begin by rewriting the $C_{t}^{(p)}(r)$ terms using the uniform / boxcar kernel $K_{\text{uniform}}(u) = \mathbf{1}_{[-1, 1]} (u)$ as
\begin{align}
	C_{t}^{(p)}(r) &= \frac{\# \left\{ \mathbf{X}^{(p)}_{t'} : \left|\left|\mathbf{X}^{(p)}_{t} - \mathbf{X}^{(p)}_{t'}\right|\right|_{\infty} \leq r\right\}}{T - p + 1} \\
	&= \frac{1}{T - p + 1} \sum_{t = 1}^{T - p + 1} K_{\text{uniform}}\left(\frac{\left|\left|\mathbf{X}^{(p)}_{t} - \mathbf{X}^{(p)}_{t'}\right|\right|_{\infty}}{r}\right) \\
	&= \frac{1}{T - p + 1} \sum_{t = 1}^{T - p + 1} \prod_{i = 0}^{p-1} K_{\text{uniform}}\left(\frac{|X_{t+i} - X_{t' + i}|}{r}\right). \label{Eqn:ApEn-as-KDE}
\end{align}
We see that (\ref{Eqn:ApEn-as-KDE}) is equivalent to the kernel density estimator for the density of $\left\{\mathbf{X}_{t}^{(p)}\right\}_{t = 1}^{T - p + 1}$ using a product of uniform kernels up to a normalization factor of $(2r)^{-p}$. The true kernel density estimator therefore would be given by
\begin{align}
	\hat{f}\left(\mathbf{x}^{(p)}\right) &= \frac{1}{T - p + 1} \sum_{t = 1}^{T - p + 1} \frac{1}{(2r)^{p}} K_{\text{uniform}}\left(\frac{||\mathbf{x}^{(p)} - \mathbf{X}_{t}^{(p)}||_{\infty}}{r}\right) \\
	&= \frac{1}{T - p + 1} \sum_{t = 1}^{T - p + 1} \prod_{i = 0}^{p-1} \frac{1}{2r}K_{\text{uniform}}\left(\frac{|x_{i} - X_{t+i}|}{r}\right). \label{Eqn:True-KDE}
\end{align}
Therefore, we see that (\ref{Eqn:cond-prob}) is the unnormalized form of (\ref{Eqn:True-KDE}) evaluated at $\mathbf{X}^{(p)}_{t}$. If we include the normalization, the summation (\ref{Eqn:def-Phi}) becomes
\begin{align}
	\Phi_{\text{normalized}}^{(p)}(r) = \frac{1}{T - p + 1}\sum_{t = 1}^{T-p+1} \log \hat{f}\left(\mathbf{X}_{t}^{(p)}\right). \label{Eqn:joint-ER}
\end{align}
If $\hat{f}$ were replaced with the true density $f$, then for large $T$, $\Phi_{\text{normalized}}^{(p)}(r)$ approximates the negative joint differential entropy
\begin{align}
	h\left[\mathbf{X}^{(p)}\right] &= -E\left[ \log f\left(\mathbf{X}^{(p)}\right) \right]\\
	&= - \int_{\mathbb{R}^{p}} f\left(\mathbf{x}^{(p)}\right) \log f\left(\mathbf{x}^{(p)}\right) \, d \mathbf{x}^{(p)}
\end{align}
by the Law of Large Numbers. However, because we evaluate the estimator $\hat{f}$ at the same data used to estimate it, (\ref{Eqn:joint-ER}) is a biased estimator of the negative differential entropy $-h\left[\mathbf{X}^{(p)}\right]$. A simple modification of (\ref{Eqn:joint-ER}), due to~\cite{kandasamy2014influence,Kandasamy:2015ux}, provides an estimator for the joint differential entropy with a fast rate of convergence in the IID case. In particular, let $\hat{f}_{-t}$ be the kernel density estimator for the joint density formed by leaving out the $t^\text{th}$ vector $\mathbf{X}_{t}$. That is, we estimate the joint density using (\ref{Eqn:True-KDE}) with all of the vectors except $\mathbf{X}_{t}$. This gives the leave-one-out (LOO) estimator for the joint differential entropy,
\begin{align}
	-\Phi_{\text{normalized, LOO}}^{(p)}(r) = -\frac{1}{T - p + 1}\sum_{t = 1}^{T-p+1} \log \hat{f}_{-t}\left(\mathbf{X}_{t}^{(p)}\right).
\end{align}
Thus we see that with the proper normalization, a modification of the Approximate Entropy gives an estimator for the finite-$p$ differential entropy rate, 
\begin{align}
	h[X_{p+1} \mid X_{p}, \ldots, X_{1}] = h[X_{1}, \ldots, X_{p+1}] - h[X_{1}, \ldots, X_{p}].
\end{align}

\bibliography{references}
\bibliographystyle{plain}

\end{document}